\documentclass[10pt,a4paper]{article}
\usepackage{f1000_styles}

\begin{document}
\pagestyle{fancy}

\newcommand{\xx}{\mathbf{X}}
\newcommand{\yx}{\mathbf{y}_{\mathbf{X}}}
\newcommand{\zz}{\mathbf{Z}}
\newcommand{\yz}{\mathbf{y}_{\mathbf{Z}}}
\newcommand{\erre}{\mathbb{R}}
\newcommand{\dd}{\mathcal{D}}
\newcommand{\xtilde}{\Tilde{x}}
\newcommand{\ddtilde}{\Tilde{\mathcal{D}}}

\newcommand{\victor}[1]{{\color{blue} #1}}
\title{An In-Depth Analysis of Data Reduction Methods for Sustainable Deep Learning}
\author[1]{Javier Perera-Lago}
\author[1]{Víctor Toscano-Durán}
\author[2]{Eduardo Paluzo-Hidalgo}
\author[1]{Rocio Gonzalez-Diaz}
\author[3]{Miguel Ángel Gutierrez-Naranjo}
\author[4]{Matteo Rucco}

\affil[1]{Department of Applied Mathematics I, University of Seville, Seville, Spain \\ \texttt{\{jperera,vtoscano,rogodi\}@us.es}}
\affil[2]{Department of Quantitative Methods, University of Loyola, Campus Sevilla, Dos Hermanas, Seville, Spain \\ \texttt{epaluzo@\{uloyola,us\}.es}}
\affil[3]{Department of Computer Science and Artificial Intelligence
University of Seville, Seville, Spain \\ \texttt{magutier@us.es}}
\affil[4]{Data Science Department, Biocentis, Milan, Italy \\ \texttt{matteo.rucco@biocentis.com}}

\maketitle
\thispagestyle{fancy}

\begin{abstract}

In recent years, Deep Learning has gained popularity for its ability to solve complex classification tasks,
increasingly delivering better results thanks to the development of more accurate models, the availability of huge volumes of data and the improved computational capabilities of modern computers. However, these improvements in performance also bring efficiency problems, 
related to the storage of datasets and models, and to the waste of energy and time involved in both the training and inference processes. In this context, data reduction can help reduce energy consumption when training a deep learning model. In this paper, we present up to eight different methods to reduce the size of a tabular training dataset, and we develop a Python package to apply 
them. We also introduce a representativeness metric based on topology to measure how similar are the reduced datasets and the full training dataset. Additionally, we develop a methodology to apply these data reduction methods to image datasets for object detection tasks. Finally, we experimentally compare how these data reduction methods affect the representativeness of the reduced dataset, the energy consumption and the predictive performance of the model.

\end{abstract}

\section*{\color{OREblue}Keywords}

Deep Learning, energy efficiency, sustainability, data reduction, dataset representativeness, classification, object detection.

\section{Introduction}

Successful Deep Learning (DL) models require considerable consumption of resources for their development, partly due to the large volumes of data used for training. As explained 
in \cite{schwartz_green_2020}, most 
Artificial Intelligence (AI) research focuses solely on improving model performance at any cost. This line of research is known as Red AI. In contrast, Green AI  
considers the energy costs associated with AI development and seeks a balance between model performance and energy efficiency.

For example, recent publications
such as \cite{zha_data-centric_2023} and \cite{xu2021survey} 
explain various ways to improve the efficiency of intelligent agents, particularly DL models. 
In \cite{verdecchia_data-centric_2022}, the authors present experiments in which a dataset is reduced by random sampling with different percentages of reduction to train various types of models, such as $k$ Nearest Neighbours ($k
$-NN), Decision Trees (DT), Support Vector Machines (SVM), Random Forests (RF), AdaBoost, and Bagging Classifier. These experiments suggest that reducing the size of the training set significantly decreases the training time in all cases and does not worsen
the model's performance in specific cases of SVM, AdaBoost, and Bagging Classifier. In the other three algorithms, random reduction significantly decreases the F1-score.

The field of object detection and localization in images has undergone a fascinating evolution over the years, driven by significant advances in computer vision and deep learning. In its early stages, the methods focused on more traditional approaches, using features and regional classifiers. However, these methods could not efficiently handle large datasets and presented challenges in terms of speed and accuracy. The introduction of Convolutional Neural Networks \cite{oshea2015introduction} (CNNs) marked a paradigm shift by addressing the automatic feature learning capability. R-CNN variants 
\cite{ren2016faster} introduced the concept of Regions of Interest (RoI), significantly improving accuracy but with a considerable computational cost.
The true milestone came with the arrival of YOLO \cite{redmon2016look}, which proposed an innovative approach by dividing the image into a grid and making predictions of bounding boxes and classes in a single pass. Although early versions of YOLO slightly sacrificed accuracy, they demonstrated revolutionary speed, making them suitable for real-time applications.
Subsequent evolution, from YOLOv2 to YOLOv5 (the model we will focus on in this analysis from now on), has been marked by continuous improvements. Later versions refined the architecture, and incorporated specialized layers and attention strategies, enhancing accuracy without significantly compromising speed.
In summary, we will focus on analyzing how data reduction affects the training of deep neural networks, expanding the list of reduction methods with other algorithms developed in recent years. The two specific tasks in which we test the selected techniques are tabular data classification and object detection for image datasets. We have tested eight different data reduction methodologies and compared their performance and efficiency in four different datasets: the Collision dataset and the Dry Bean dataset, both containing tabular data focused on classification, and the Roboflow dataset and the Mobility Aid dataset, which consist of images and focused on object detection (localizing people in wheelc\-hairs and pedestrians). Besides, we propose a specific methodology tailored for dealing with images, which is particularly crucial for detecting people in wheelchairs and pedestrians in the REXASI-PRO context. Finally, we have unified the Python implementations into two user-friendly GitHub repositories: one serving as a library for reduction methods and the other showcasing the experiments carried out, as cited in \cite{Perera_Lago_Repository_Survey_Green_2023} and \cite{Victor_Toscano_Duran_Repository_experiments_Survey_2023}.

This paper is organized as follows: In Section \ref{sec:keyFindingsContributions}, we present the Python library created to utilize data reduction methods, the proposed methodology for applying them to structured data, such as images, and a summary of the main results obtained and conclusions. Later, in Section \ref{sec:background}, all the preliminary concepts are introduced, including key concepts about multi-layer perceptrons, how the classification of tabular data and object detection from images works, along with metrics to measure the performance of data reduction methods. Moving on to Section \ref{sec:data_reduction_methods}, we introduce different methodologies to reduce a dataset, categorizing them into four groups: statistic-based methods, geometry-based methods, ranking-based methods and wrapper methods. 
Finally, in Section \ref{sec:experiments}, the results are tested experimentally for both tabular data and object detection, providing details about the databases, the experiment setup and the parameter settings.

\section{Key Findings and Contributions}\label{sec:keyFindingsContributions}

In this section, we highlight the main contributions developed for this paper.
    
\subsection{Python Library: Data Reduction Methods}
    
We have released the Beta version of a Python Library to use the data reduction methods. It is placed on an Open-access GitHub repository provided with instructions from the \texttt{README.md} file of the repository to use the data reduction methods. 

\begin{center}
\begin{tabular}{|c|c|}
\hline
   Library repository  & \cite{Perera_Lago_Repository_Survey_Green_2023} \\\hline
   Experiments repository  & \cite{Victor_Toscano_Duran_Repository_experiments_Survey_2023}\\\hline
\end{tabular}    
\end{center}

The reduction methods implemented are listed in Table~\ref{tab:DataReductionMethods}. We have included those 
state-of-the-art methods for date reduction that satisfy the following conditions: \begin{enumerate}
\item The implementation is available or their implementation is straightforward.
\item The final size of the reduced dataset can be chosen. 
\item The reduction time offsets the training savings tested in small examples.
\end{enumerate}

\begin{table}[h]
\centering
\begin{tabular}{|l|l|l|l|}
\hline
\textbf{STATISTIC-BASED} & SRS & Stratified Random Sampling & \cite{verdecchia_data-centric_2022} \\ \cline{2-4}
& PRD & ProtoDash & \cite{gurumoorthy2019efficient} \\ \hline
\textbf{GEOMETRY-BASED} & CLC & Clustering Centroids & \cite{olvera2010review} \\ \cline{2-4}
& MMS & Maxmin Selection & \cite{lacombe_data-driven_2021} \\ \cline{2-4}
& DES & Distance-Entropy Selection & \cite{li_distance-entropy_2022} \\ \hline
\textbf{RANKING-BASED} & PHL & PH Landmarks & \cite{stolz2023outlier} \\ \cline{2-4}
& NRMD & Numerosity Reduction by Matrix Decomposition & \cite{ghojogh2019instance} \\ \hline
\textbf{WRAPPER} & FES & Forgetting Events Score & \cite{toneva2018empirical} \\ \hline
\end{tabular}
\caption[Data Reduction Methods]{List of state-of-the-art data reduction methods selected for comparison.}
\label{tab:DataReductionMethods}
\end{table}
    
\subsection{Data Reduction for Images}\label{sec:drImages}

Some of the methods listed in Table~\ref{tab:DataReductionMethods} need their input to be $n$-dimensional vectors and cannot be directly applied to structured data such as images that have a tensor of shape (height, width, channels). 
In this paper, we propose two methodologies (see Section~\ref{sec:methodology}) to extend the reduction methods to images, in the context of this research project (object detection). Specifically,

\begin{enumerate}
    \item For statistic-based, geometry-based, and ranking-based reduction methods: The proposed me\-thodology involves using a feature extraction model, such as the YOLOv5 backbone, and applying it to the given dataset. Then, we employ Global Average Pooling (GAP) \cite{gholamalinezhad2020pooling} to transform the images into $n$-dimensional vectors (768 dimensions, due to the backbone's structure, which yields 768 feature maps). This allows us to apply data reduction techniques effectively. Finally, the images selected by the reduction method as the most important ones will be used to train the model. 
    \item For the wrapper method: We create a classification network to apply these techniques internally within the classification model. Subsequently, the images selected as the most important by the reduction method will be used to tune the YOLOv5 model with the reduced dataset.
\end{enumerate}

\subsection{Summary of the Main Results and Conclusions}

The outcome of the investigation on the effectiveness of reduction methods to achieve green AI, with the aim of minimizing $\text{CO}_2$ emissions and computational costs while maintaining performance, yielded promising results. The CO2 emissions were estimated using a specific software implementation but no physical sensors were used (See Section~\ref{sec:efficiency}).

\begin{itemize}
    \item For tabular datasets: We have found that using reduced datasets notably decreases both the computation time and carbon emissions of neural network training. We have also found that these reduction methods can discard a large number of training examples without losing the good predictive properties of the DL models. However, notice there is no reduction method that always performs better than the rest. Furthermore, for one of the analyzed datasets, we have found a significant statistical correlation between the performance metrics of the trained models and a topological metric called $\varepsilon$-representativeness  (discussed in detail in Section \ref{sec:topologicalMeasure}), which measures how close is the reduced dataset to the full one.
    \item For images: The proposed methodology is an effective way to achieve greener artificial intelligence models with good performance. 
    In summary, regarding the Roboflow dataset, 
    substantial reductions in both $\text{CO}_2$ emissions and computation time (particularly for a 75\% reduction rate) were achieved (approximately 60\% time savings) 
    without compromising the model's performance in localization and object detection tasks.
    Among the reduction methods applied, SRS, MMS and RKM proved most effective, while NRMD exhibited poorer performance, and FES incurred longer processing times. Similarly, positive results were observed with the Mobility Aid dataset, with significant reductions in $\text{CO}_2$ emissions and computation time without compromising performance for specific methods. In this case, SRS, PRD, PHL and MMS demonstrated superior effectiveness, while NRMD and RKM were identified as less efficient methods.
\end{itemize}

\section{Background}\label{sec:background}

In this section, all preliminary concepts are introduced. We provide key concepts about multi-layer perceptrons and present the two specific problems we are addressing: the classification of tabular data and object detection from images. Subsequently, we describe the YOLOv5 architecture. 
Finally, in Section~\ref{sec:metrics}, we discuss the different metrics to measure the performance of the different data reduction methods.

\subsection{Multi-layer Perceptrons}

Multi-layer perceptrons (MLP) are the simplest kind of neural networks~\cite{Surdeanu_Valenzuela-Escárcega_2024}. 
An MLP, denoted $\mathcal{N}$, is a function that transforms input vectors through a series of smaller functions called layers. Formally, $\mathcal{N}$ can be represented as a composition $\mathcal{N} = f_l \circ \cdots \circ f_1$, where $f_j:\erre^{d_{j-1}}\rightarrow\erre^{d_j}$ is the $j$-th layer function. 
Each layer $f_j$ can be decomposed into $d_j$ smaller functions called units or neurons, $f_j = (f_{j,1}, \ldots, f_{j,d_j})$. 
Each neuron $f_{j,m}$ is defined as $f_{j,m}(x) = g_{j}(W_{j,m}^Tx + b_{j,m})$, where $W_{j,m} \in \mathbb{R}^{d_{j-1}}$ is the weight vector of the neuron, $b_{j,m} \in \mathbb{R}$ is the bias term, and $g_{j}: \mathbb{R} \rightarrow \mathbb{R}$ is a nonlinear activation function \cite{Agostinelli2014LearningAF}. These components determine how information flows through the network and how it is transformed at each layer. 

The design of an MLP depends on three elements: its architecture, a vector of parameters, and a set of learning hyperparameters. The architecture of an MLP is the choice for the number of layers ($l$), the number of dimensions of the layers ($d_1, ..., d_l$) and the activation functions ($g_1, g_2, ..., g_l$). Given a neural architecture, the network parameters are the weight vector entries $W_{j,m}$ and the bias terms $b_{j,m}$. All these parameters can be encoded in a parameter vector $\theta \in \erre^p$, being $p$ the number of adjustable parameters in the network. It is common to denote the MLP by $\mathcal{N}_{\theta}$ to state that $\mathcal{N}$ uses $\theta$ as a parameter vector. 
The set of all possible parameter vectors is denoted as $\Theta$.
Finally, the hyperparameters are related to the training procedure, that is, the search for a vector $\theta$ that makes $\mathcal{N}_{\theta}$ useful for the specific task and dataset. The hyperparameters must be defined prior to this search and are usually set by trial and error.

Given an MLP $\mathcal{N}_{\theta}$ and a tabular dataset $\dd$ (whose definition can be read in Subsection \ref{subsec:tabulardata}), the suitability of the task is measured by a loss function $\mathcal{L}(\cdot,\dd): \Theta \rightarrow \erre$ \cite{wang2020comprehensive}, designed to have small values when $\mathcal{N}_{\theta}$ is useful and vice versa. 
Two typical examples of loss functions are the Mean Square Error for regression tasks and the Categorical Cross Entropy loss for classification tasks \cite{Mao2023CrossEntropyLF}. 
The search for a good choice of $\theta$ is made with an iterative training process that minimizes $\mathcal{L}$ across $n_e$ successive epochs using the information from $\dd$. 
In each epoch, $\dd$ is randomly partitioned into a series of sub-datasets $\mathcal{B}_1, \mathcal{B}_2, \ldots$, known as batches, with a maximum size equal to $\beta$. 
Each batch $\mathcal{B}_j$ is then fed into $\mathcal{N}_{\theta}$, where its performance is evaluated, resulting in the calculation of the loss gradient $\nabla_{\theta}\mathcal{L}(\theta, \mathcal{B}_j)$. 
This gradient information is used by an optimization algorithm such as Stochastic Gradient Descent (SGD), Adaptative Moment Estimation (Adam) or Root Mean Square Propagation (RMSprop), to update $\theta$ and minimize $\mathcal{L}$ \cite{Ruder2016AnOO}. 
These optimization algorithms may depend on other hyperparameters such as a learning rate $\gamma \in \erre^+$ or a momentum $\mu \in \erre^+$. 
Additionally, one can complement the training process with regularization techniques such as L1 or L2 regularization, dropout, or weight decay, which try to prevent overfitting (which occurs when the model fits so closely to the training dataset that it does not generalize well to new examples). 
The hyperparameters that define the training process are then the loss function $\mathcal{L}$, the number of epochs $n_e$, the batch size $\beta$, the optimization algorithm (together with other associated hyperparameters if needed), and the regularization techniques. For more details on feedforward neural networks, MLPs and how to train them, visit \cite{islam2019overview}.

\subsection{Classification of Tabular Data}
\label{subsec:tabulardata}

A dataset $\dd=(X,f)$ of a classification problem is a pair composed of a set of examples $X = \{x_1,\cdots x_N\}\subset \mathbb{R}^d$ and a function $f:X \rightarrow\{1,\dots, c\}$. 
The function $p_j:X \rightarrow \erre$ that maps $x_i=(x_{i,1},\cdots,x_{i,d})^T$ to $p_j(x_i)=x_{i,j}$ is called the feature $j$. The set of examples $X_k = \{x_i\in X: f(x_i)=k\}$ with $k\in\{1,\dots,c\}$ are called the class $k$. 
These types of datasets are usually known as tabular data because they can also be seen as a pair $(\mathbf{X},\mathbf{y})$ where $\mathbf{X} \in \erre^{N\times d}$ is a table or matrix whose $i$-th row corresponds to the example $x_i$ and whose $j$-th column corresponds to the feature $p_j$, and $\mathbf{y} \in \erre^n$ is a vector whose $i$-th component corresponds to $y_i=f(x_i)$. 
This tabular representation of the dataset is usual in computer science and is how it is used in Python code to design DL models. 
Given a dataset $\dd=(X,f)$, the classification problem consists of finding a feed-forward neural network $\mathcal{N}:\mathbb{R}^d \rightarrow \{1,\dots, c\}$ that approximates $f$.

\subsubsection{Data Reduction}

Given a dataset $\dd=(X,f)$, the goal of data reduction is to find a reduced dataset $\dd_R=(S,g)$, where $S = \{s_1,\cdots s_n\}\subset \mathbb{R}^d$ (being $n<N$) and $g:S \rightarrow\{1,\dots, c\}$. 
The class $k$ in $\dd_R$ will be denoted as $S_k = \{s_i\in S: g(s_i)=k\}$. 
In Section \ref{sec:data_reduction_methods}, some algorithms to extract a reduced dataset $\dd_R$ from $\dd$ are described. 
In most of them, the resulting $\dd_R$ will be a sub-dataset of $\dd$, which means that $S \subset X$ and $g = f|_{S}$. 
Assuming that $\dd_R$ is representative of $\dd$ and inherits its intrinsic properties, it should be able to be used to train $\mathcal{N}$ instead of $\dd$, gaining efficiency and obtaining a model $\mathcal{N}_R$  with similar performance. 
To test the correlation between representativeness, efficiency and performance it is necessary to define adequate metrics for all of them. 
The metrics that we use in our experiments can be consulted in Subsection \ref{sec:metrics}.
   
\subsection{Object Detection from Images}

This Subsection is devoted to the definition of the problem and the description of YOLOv5.

\subsubsection{Defining Object Detection}

The problem of object detection \cite{zaidi2021survey} and localization in the field of computer vision refers to the task of identifying the presence of specific objects in an image and providing accurate information on the spatial location of each of them. In other words, its main goal is to detect the presence of objects of interest within a visual scene and, at the same time, delineate the exact region where they are located in the image.

        To better understand this type of problem, it can be helpful to break it down into two key components:
        
        \begin{itemize}
        \item Object Detection: This involves identifying and classifying the presence of objects within an image. This aspect addresses the fundamental question of "What objects are in the image?". Each detected object is usually associated with a specific class.
        \item Object Location: Refers to providing information about the precise spatial location of the detected objects. This involves defining the coordinates or bounding boxes surrounding each object in the image, indicating exactly where they are located.
        \end{itemize}

        \subsubsection{Object Detection with YOLOv5}

        YOLO \cite{redmon2016look}, an abbreviation for "You Only Look Once", is conceived to address object detection with a comprehensive and efficient approach. 
        Its main objective is to perform accurate detections of multiple objects in a single pass through the image, minimizing duplication of efforts and optimizing processing speed. 
        YOLO is a single-stage architecture with which object detection is performed by viewing the problem as a regression problem to spatially separate the bounding box and the probability classes associated with the bounding box. 
        A neural network predicts the bounding box and prediction class directly from the entire image from a single evaluation.
        The fifth version of YOLO, named YOLOv5 \cite{yolov5Ultralytics, jacob2020yolov5}, is the first native release of models in the YOLO family written in Pytorch \cite{Pytorch}. 
        YOLOv5 is fast, with inference times up to 0.007 seconds per image, meaning 140 frames per second. Figure \ref{image-YOLO} shows the detailed architecture of YOLOv5. Specifically, YOLOv5 consists of three essential components:
        
        \subsubsection*{Backbone}  
        
        The backbone extracts the essential features from the input image. 
        In YOLOv5, it includes CSP-Darknet53, which is a convolutional neural network and incorporates a cross-stage partial network (CSPNet) \cite{wang2019cspnet} into Darknet to separate the base layer feature map into two parts and then combine them through a cross-stage hierarchy as shown in Fig. \ref{image-YOLO}. X is a variable in CSP1\_X and CSP2\_X, meaning the number of BottleNecks in the network. This enhanced CSPNet, built upon Darknet53, residual blocks, depthwise separable convolutions, and preactivation for improved efficiency and feature representation. For instance, given an RGB input image of 416 pixels in height and width, the backbone produces an output with 768 feature maps, each with dimensions of 13 pixels in height and width.

        \subsubsection*{Neck}
        
\begin{figure}
        \centering
    \includegraphics[width = 0.7\textwidth]
{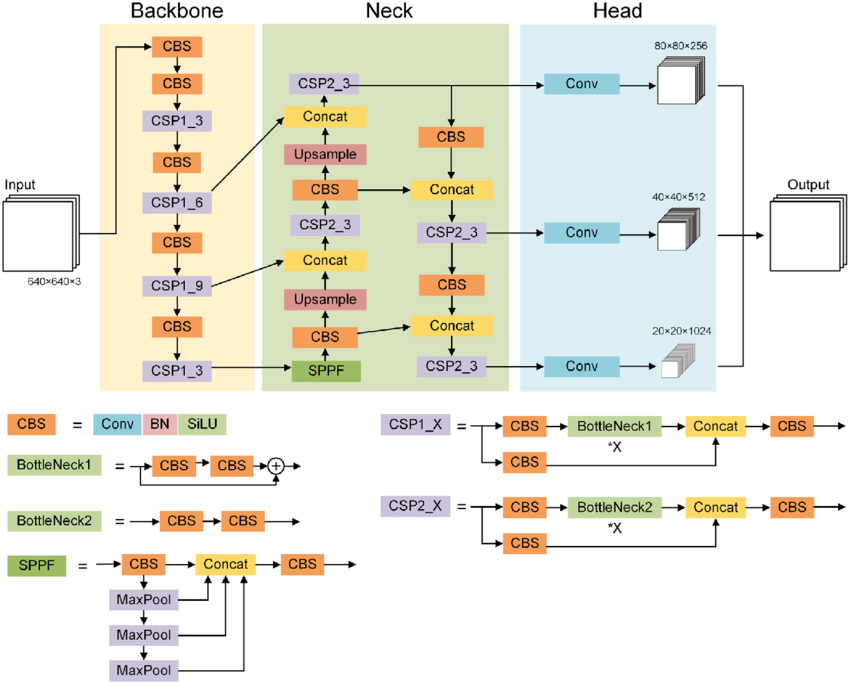}
        \caption[YOLOv5 Architecture]{\label{image-YOLO} Architecture of YOLOv5 \cite{YOLOV5Arquitectura}, including three main parts: backbone, neck and head. The "backbone" is responsible for extracting fundamental features from the image, such as edges and textures. The "neck" is used to extract feature pyramids, which helps the model to generalize well to objects of different sizes and scales. Finally, the "head" is responsible for the final prediction, generating the coordinates and classes of the detected objects.}
        \end{figure}

        The neck acts as a bridge between the backbone and the head, performing operations to merge and refine features at different scales.  
        The neck includes a spatial pyramid pooling-fast (SPPF) layer and a cross-stage partial path aggregation network (CSP-PAN), as shown in Fig. \ref{image-YOLO}.
        A spatial pyramid pooling (SPP) \cite{He_2014} layer is a pooling layer that removes the CNN limitation of fixed-size input images. 
        The SPPF layer optimizes the SPP structure and improves the efficiency more than twofold. It aggregates information received from inputs and returns a fixed-length output. 
        PAN \cite{liu2018path} is a feature pyramid network, used to improve information flow and help with the proper location of pixels in mask prediction task. In YOLOv5, this network has been modified applying the CSPNet strategy as shown in Fig.~\ref{image-YOLO}.
        
        \subsubsection*{Head}
        
        The network's head makes the final predictions, generating bounding boxes and classifications for each object. It is composed of four convolution layers that predict the location of the bounding boxes (x,y,height,width), the scores and the final classification.
        In addition, YOLOv5 uses several augmentations such as Mosaic, copy-paste, random affine, MixUp, HSV augmentation, random horizontal flip, as well as other augmentations from the \texttt{albumentations} package \cite{info11020125}. 
        It also improves the grid sensitivity to make it more stable to runaway gradients.

        YOLOv5 provides five scaled versions: YOLOv5n (nano), YOLOv5s (small), YOLO\-v5m (medium), YOLO\-v5l (large), and YOLOv5x (extra large), where the width and depth of the convolution modules vary depending on the specific applications and hardware requirements. From now on, we will focus on YOLOv5m, which is the one we have used for the experimentation.

\subsection{Metrics}\label{sec:metrics}

In the following, we will discuss metrics for evaluating the performance of artificial intelligence models for both classification and object detection, as well as metrics for evaluating the representativeness of the reduced data over the entire dataset and evaluating the cost of the models.

\subsubsection{Representativeness Metrics}\label{sec:topologicalMeasure}

In this section, we include the two metrics that we will use in our experiments to measure the similarity between the original dataset $\dd$ and its reduced version $\dd_R$. 

\subsubsection*{Reduction Ratio}

This is a common metric in the machine learning literature for comparing $\dd$ and $\dd_R$. The reduction ratio is just the quotient between the sizes of $X_R$ and $X$. It has already been considered in publications such as \cite{verdecchia_data-centric_2022} and \cite{ghojogh2019instance} to measure the representativeness of $\dd_R$ with respect to $\dd$ and to study its effect on efficiency and performance, and we will also use it for our experiments.

\subsubsection*{$\varepsilon$-Representativeness}

In \cite{gonzalez-diaz_topology-based_2022}, the authors introduced the concept of $\varepsilon$-representative datasets. The $\varepsilon$ indicates how representative and, hence, how good is the representation of $\dd_R$ for $\dd$, smaller values being better. Consequently, we have $\varepsilon=0$ if and only if $\dd_R=\dd$.
Given a fixed isometry $i:\dd_R \rightarrow \mathbb{R}^d$, the minimum $\varepsilon$ such that $\dd_R$ is a $\varepsilon$-representative dataset of $\dd$ is:

\begin{equation}
\label{eq:min_epsilon}
\varepsilon^* = \max _{k=1,\cdots,c} \; \max_{x \in f^{-1}(k)} \; \min_{x' \in f_R^{-1}(k)} ||x-i(x')||
\end{equation}

It has been mathematically proven in \cite{gonzalez-diaz_topology-based_2022} that, if $\dd_R$ is $\varepsilon$-representative of $\dd$ with $\varepsilon$ small enough, $\dd_R$ and $\dd$ have the same accuracy (a performance metric whose definition can be read in \ref{subsubsec:accuracy}) for a perceptron (which is a feed-forward neural network $\mathcal{N}:\erre^d \rightarrow \erre$ with a single layer). Besides, it was proved experimentally in \cite{gonzalez-diaz_topology-based_2022} that a similar relationship exists between the $\varepsilon$-representativeness of $\dd_R$ with respect to $\dd$ and the model performance for more complex neural architectures.

\subsubsection{Performance Metrics}

Given a dataset $\dd=(X,f)$ of a classification problem, one can design many different DL models with different architectures and parameters, that will give different approximations to $f$. 
It is then essential to measure how good these approximations are, assessing the overall performance of each model. 
Depending on the specific goals and characteristics of the classification task, some metrics might be more relevant than others, and the final choice of metrics is up to the model developers.

All the performance metrics that we will use in our experiments are derived from the confusion matrix. 
Given a dataset $\dd$ and a DL model $\mathcal{N}$, the confusion matrix is a table with $c$ rows and $c$ columns where the cell $(i,j)$ is filled with a non-negative integer $n_{i,j}$ equal to the number of examples in $X$ whose actual class is $f(x)=i$ and whose predicted class is $\mathcal{N}(x)=j$. The sum of all the entries in row $i$ is equal to the number of examples whose actual class is $i$, and it is denoted as $A_i$. The sum of all the entries in column $j$ is equal to the number of examples whose predicted class is $j$, and is denoted as $P_j$. 
In addition, the sum of all the numbers in the confusion matrix is equal to $N$, the size of $X$. 
An example of a confusion matrix can be found in Table \ref{tab:confusionmatrix}. 
Ideally, one would like to obtain a DL model $\mathcal{N}$ that fits exactly $f$, that is, $f(x)=\mathcal{N}(x) \ \forall x \in X$. 
In that case, the confusion matrix would be null out of the diagonal. 
In practice, it is not always possible to find such a perfect model (it may not be desirable due to the risk of overfitting $X$) but, in general, it is considered a good sign to get a confusion matrix with high values in the diagonal entries and lower values in the non-diagonal entries.

\begin{table}
\hrule height 0.05cm  \vspace{0.1cm}
\centering
\begin{tabular}{|c|c|c|c|c|c|c|} 
\hline
\cellcolor{gray!20} \textbf{} & \cellcolor{gray!20} \textbf{Predicted 1} & \cellcolor{gray!20} $\cdots$ & \cellcolor{gray!20} \textbf{Predicted k} & \cellcolor{gray!20} $\cdots$ & \cellcolor{gray!20} \textbf{Predicted c} & \cellcolor{gray!20} \textbf{Total} \\
\hline
\cellcolor{gray!20} \textbf{Actual 1} &  $n_{1,1}$ &  $\cdots$ &  $n_{1,k}$ &  $\cdots$ &  $n_{1,c}$ &  $A_1$ \\
\hline
\cellcolor{gray!20} $\vdots$ &  $\vdots$ &  $\ddots$ &  $\vdots$ &  $\ddots$ &  $\vdots$ &  $\vdots$ \\
\hline
\cellcolor{gray!20} \textbf{Actual k} &  $n_{k,1}$ &  $\cdots$ &  $n_{k,k}$ &  $\cdots$ &  $n_{k,c}$ &  $A_k$ \\
\hline
\cellcolor{gray!20} $\vdots$ &  $\vdots$ &  $\ddots$ &  $\vdots$ &  $\ddots$ &  $\vdots$ &  $\vdots$ \\
\hline
\cellcolor{gray!20} \textbf{Actual c} &  $n_{c,1}$ &  $\cdots$ &  $n_{c,k}$ &  $\cdots$ &  $n_{c,c}$ &  $A_c$ \\
\hline
\cellcolor{gray!20} \textbf{Total} &  $P_1$ &  $\cdots$ &  $P_k$ &  $\cdots$ &  $P_c$ &  $N$ \\
\hline
\end{tabular}
 \caption[Scheme of the Confusion Matrix for a Classification Problem with $c$ Classes]{\label{tab:confusionmatrix}Confusion matrix for a classification problem with $c$ classes, together with its marginal sums.}
\end{table}

\subsubsection*{Accuracy}
\label{subsubsec:accuracy}

Accuracy is the most straightforward performance metric. It is the probability of correctly classifying a random example from $X$ using $\mathcal{N}$. It is calculated as the quotient between the number of correctly classified examples and the total size of $X$, that is, 

\begin{equation}
\label{eq:acc}
Acc = \frac{\sum_{k=1}^{c}n_{k,k}}{N}
\end{equation}

Accuracy is a good metric for getting an overview of the model quality, but it can be misleading if the evaluation is reduced to it.
When the training dataset is imbalanced (that is, it has some classes much more numerous than others) it is possible to find DL models that are bad at classifying the items from the least populated classes but yet have high accuracy because they perfectly fit $f$ for the most populated ones. For this reason, it is also necessary to use other performance metrics that analyze the performance of $\mathcal{N}$ class by class.

\subsubsection*{Precision}

Precision measures the probability that the model $\mathcal{N}$ is correct when it predicts that an example belongs to a specific class $k$. It tells us how confident we can be in the predictions obtained. For each class $k$, the precision is obtained as the quotient between the number of examples correctly classified in class $k$ and the total number of examples whose predicted class is $k$, that is,

\begin{equation}
\label{eq:prek}
Pre_k = \frac{n_{k,k}}{P_k}
\end{equation}

When working with a dataset $\dd$ with $c$ classes, we can calculate $c$ different precision values from the confusion matrix. We can resume all this information using the macro average precision, which is just the arithmetic mean of all of them:

\begin{equation}
\label{eq:maprek}
MAPre = \frac{1}{c}\cdot \sum_{k=1}^{c} Pre_k
\end{equation}

The macro average precision assigns the same relevance to the precision of all classes regardless of their size. It also mitigates the bias induced by the largest ones.

\subsubsection*{Recall}

"Recall" measures the probability that the model $\mathcal{N}$ correctly classifies the examples from $X_k$. It tells us how good the predictions are for that specific subset of $X$. For each class $k$ the recall is obtained as the quotient between the number of examples correctly classified in class $k$ and the total number of examples whose actual class is $k$, that is,

\begin{equation}
\label{eq:reck}
Rec_k = \frac{n_{k,k}}{A_k}
\end{equation}

Just as all the precisions can be summarized with the macro average precision, the recall values can also be aggregated using the macro average recall, with the analogous formula:

\begin{equation}
\label{eq:mareck}
MARec = \frac{1}{c}\cdot \sum_{k=1}^{c} Rec_k
\end{equation}
Some bibliographic sources also refer to this metric as Balanced Accuracy \cite{brodersen2010balanced}.

In object detection, a high Recall means the model is proficient at capturing most of the objects in the images, minimizing the number of false negatives.

\subsubsection*{F1-score}

F1-score is a metric that gives us a trade-off between precision and recall. For each class $k$, it is obtained as the harmonic mean of the precision and recall values:

\begin{equation}
\label{eq:f1k}
F1_k = 2 \cdot \frac{Pre_k\cdot Rec_k}{Pre_k+Rec_k} = 2 \cdot \frac{n_{k,k}}{P_k+A_k}
\end{equation}

As a harmonic mean, $F1_k$ lies between $Pre_k$ and $Rec_k$, always being less than or equal to the arithmetic mean $\frac{Pre_k+Rec_k}{2}$. 
In fact, $F1_k$ tends to approach the minimum value between $Pre_k$ and $Rec_k$, resulting in a lower score when precision or recall is low. 
Maximizing the F1-score is desirable because achieving a high value indicates that both precision and recall are high, proving that the model $\mathcal{N}$ has a good performance for that specific class.

As we already did with precision and recall, we can define the macro average F1-score as:

\begin{equation}
\label{eq:maf1k}
MAF1 = \frac{1}{c}\cdot \sum_{k=1}^{c} F1_k
\end{equation}

An alternative definition for the macro average F1-score can be found in \cite{sokolova2009systematic}, but we prefer to use this one since it appears to be more suitable \cite{opitz2019macro} and can be computed with standard Python libraries devoted to classification learning.

\subsubsection*{Intersection over Union and Mean Average Precision} 

Intersection over Union ($IoU$) \cite{rezatofighi2019iou} indicates the overlap of the predicted bounding box coordinates with the ground truth box. Higher $IoU$ indicates that the predicted bounding box coordinates closely resemble the ground truth box coordinates. $IoU$  is calculated by comparing the overlapped region between the prediction of the model and the ground truth with the total region covered by both. Mathematically, $IoU$ is expressed as the ratio of the intersection area to the union area of the two regions:
\begin{equation}
\label{eq:IoU}
IoU = \frac{Intersection Area}{Union Area}\,,
\end{equation}
        where:
\newline\noindent        
        - Intersection Area: Area where the model's prediction and the ground truth overlap.
    \newline\noindent        
        - Union Area: Total area covered by both regions.

$IoU$ is commonly set at 0.5, which means that the metric considers a detection successful if at least 50\% of the predicted region overlaps with the actual region of the object.

The Mean Average Precision ($mAP$) \cite{henderson2017map, deval2022map} is the current benchmark metric used by the computer vision research community to evaluate the robustness of object detection models. The $mAP$ metric evaluates the overall accuracy of the model across multiple intersections over Union thresholds, so the first thing you need to do when calculating the Mean Average Precision ($mAP$) is to select the $IoU$ threshold. When calculating $mAP$, you have the flexibility to choose either a single $IoU$ threshold or a range of thresholds. For instance, when you choose a single IoU threshold, such as 0.5 (denoted $mAP$@0.5), you are assessing the model's accuracy when the predicted bounding box overlaps with the ground truth bounding box by at least 50\%. However, seting a range of thresholds, like 0.5 to 0.95 with 0.05 increments (indicated as $mAP$@0.5:0.95), allows you to evaluate the model performance across a spectrum of IoU values.

The second thing we need to do is divide our detections into classes based on the detected class. We then compute the Average Precision ($AP$) for each class and calculate its mean, resulting in an $mAP$ for a given $IoU$ threshold. We calculate Precision-Recall curve points for different confidence thresholds and then we calculate the $AP$ for each class $k$ using the following equation:

\begin{equation}
\label{eq:APk}
    AP_k = \frac{1}{t}\int_0^1 Pre_k(Rec_k)d Rec_k
\end{equation}
where $t$ is the number of $IoU$ thresholds considered.

All the average precisions can be aggregated using the mean Average Precision, by the following formula:

\begin{equation}
\label{eq:mAP}
   mAP= \frac{1}{c}\cdot \sum_{k=1}^{c} AP_k
\end{equation}

The $mAP$ incorporates the trade-off between precision and recall and considers both false positives (FP) and false negatives (FN). This property makes $mAP$
a suitable metric for most detection applications. A high $mAP$ means that a model has both a low false negative and a low false positive rate. 

\subsubsection{Efficiency Metrics}\label{sec:efficiency}

The cost of designing and using a model can depend on many factors. According to 
\cite{schwartz_green_2020}, the total cost of getting a result (R) is linearly related to the cost of processing a single example (E), the size of the training dataset (D) and the number of hyperparameters to be set (H), giving us the following equation:

\begin{equation}
\label{eq:cost}
    \operatorname{Cost}(R) \propto E \cdot D \cdot H
\end{equation}

Following this idea, several metrics have been proposed to measure the amount of work performed during training, such as electricity consumption, the number of parameters to be adjusted, or the total number of floating point operations performed. 
Some advantages and disadvantages of using these metrics can be read in \cite{schwartz_green_2020}. For our experiments, we will only focus on two specific metrics that are easy to calculate with Python code and give us an intuition about the impact that our DL model has on the environment. These are the elapsed computing time needed to build the DL model and the estimated carbon emission into the atmosphere during the process.

\subsubsection*{Elapsed Computing Time}

Measuring the elapsed computation time is as simple as setting a timer at the beginning of the model building and using it to know how many seconds have passed until the whole process is finished. This time span can be influenced by factors independent of the training dataset and the model, such as hardware specifications, concurrent tasks on the same machine, and the utilization of multiple cores. However, it serves as a natural metric. When these factors are kept constant, the computation time serves as a direct indicator of energy consumption and carbon emissions, making it a meaningful measure of efficiency in the model-building process.

\subsubsection*{Estimated Carbon Emission}

Carbon emission is the quantity we want to minimize, since carbon dioxide ($\text{CO}_2$) is one of the main gases involved in the greenhouse effect. An excessive release of $\text{CO}_2$ into the atmosphere contributes to the change in its composition, leading to an increase in the global average temperature \cite{stocker2013technical, myhre2014anthropogenic}. Nevertheless, in practice, it is not easy to give an exact measure of carbon emission, since it depends on the sources of the energy used, the computer where the calculations are done, and the quality of the local electricity infrastructure. However, it is possible to give an approximate measure of carbon emission using the Python package CodeCarbon \cite{codecarbon}. 

This approximation measure is the product of the energy consumed by its carbon intensity, which is the amount of $\text{CO}_2$ released per unit of energy. The amount of energy consumed by a computer is estimated by monitoring the power usage of its components, such as the central processing unit (CPU), graphics processing unit (GPU), and random access memory (RAM). To obtain the carbon intensity of the energy, it is necessary to know where it comes from. Each energy source emits a different amount of $\text{CO}_2$ for each kilowatt-hour of energy generated. Coal, petroleum and natural gas are three sources with high carbon intensity, while renewable sources such as solar power and hydroelectricity are characterized by lower carbon intensity. With the combination of energy sources used in the geographical area where the computer is located (the so-called energy mix), the average carbon intensity can be computed. This methodology is based on \cite{lottick2019energy} and has already been used to estimate the carbon emission of machine learning development in \cite{verdecchia_data-centric_2022}.
        
\section{Data Reduction Methods}
\label{sec:data_reduction_methods}

In this section, we introduce different methodologies to reduce a dataset. According to the nature of the reduction algorithm, we categorize these methods into four groups: statistic-based methods, which reduce the dataset using probability or statistical concepts; geometry-based methods, which take into account the distances between examples to reduce the dataset; ranking-based methods, which sort the examples according to some criterion and reduce the dataset by selecting the best ones; and wrapper methods, which reduce the dataset during the training process. 

\subsection{Statistic-based Methods}

In this subsection, we introduce two data reduction methods that use concepts of probability and statistics to extract a reduced dataset $\dd_R$ from $\dd$.

\subsubsection*{Stratified Random Sampling (SRS)}

The most simple method for data reduction is Stratified Random Sampling (SRS), as proposed in \cite{verdecchia_data-centric_2022}, where the natural strata are the $c$ classes of $\dd$. Given a proportion $p \in [0,1]$, the algorithm just selects for each class $k$ a random subset $S_k \subset X_k$ with a reduction ratio of $p$. This ensures that $\dd_R$ has the same class balance as $\dd$. The pseudocode for SRS is shown in Algorithm \ref{alg:SRS}.

\begin{algorithm}
\caption{SRS: Stratified Random Sampling}
\label{alg:SRS}
\KwData{$\dd=(X,f)$, \ $p\in [0,1]$}
\KwResult{$\dd_R=(S,g)$}
\For{$k=1,\cdots,c$}{
Set the class $k$ as $X_k=\{x \in X: f(x)=k\}$\;
Set the number of examples to be selected as $n_k = \lfloor p \cdot |X_k| \rfloor$\;
Select a random subset $S_k \subset X_k$ with $|S_k|=n_k$ \;
}
Set $S=\bigcup_{k=1,\cdots,c}S_k$\;
Set $g = f|_{S}$\;
\end{algorithm}

\subsubsection*{ProtoDash Selection (PRD)}

ProtoDash Selection (PRD) \cite{gurumoorthy2019efficient} is an algorithm based on the concept of Maximum Mean Discrepancy (MMD), which measures the dissimilarity between two probability distributions by comparing finite samples. Given a set of indices $I = \{1,\cdots,n_A\}$, let  $A = \{a_i \in \erre^d\}_{i=1}^{n_A}$ be the sample. Given a subset of indices $L \subset I$, a vector of non-negative weights $w = (w_1,\cdots,w_{n_A})^T$ with $w_j = 0 \ \forall j \not \in L$, and a kernel function $K: \erre^d \times \erre^d \rightarrow \erre_+$, the empirical maximum mean discrepancy between $A$ and $B = \{a_j:j \in L\}$ is: 

\begin{equation}
\label{eq:MMD}
\widehat{MMD}(K,A,B,w) = \frac{1}{n_A^2}\sum_{i,j \in I} K(a_i,a_j) - \frac{2}{n_A}\sum_{j\in L}w_j\sum_{i\in I}K(a_i,a_j) + \sum_{i,j \in L}w_iw_j 
K(a_i,a_j)
\end{equation}

The aim of ProtoDash Explainer is to find a subset $L \subset I$ with size $|L|=m$ and a vector of weights $w = (w_1,\cdots,w_{n_A})^T$ that minimize $\widehat{MMD}(K,A,B,w)$, which is equivalent to maximize Eq.~(\ref{eq:lw}).

\begin{equation}
\label{eq:lw}
l(w) = w^T\mu - \frac{1}{2} w^T K w
\end{equation}
being $\mu_{j} = \frac{1}{n_A}\sum_{i}K(a_i,a_j)$ the $j$-th component of the vector $\mu$ and $K_{i,j} = K(a_i,a_j)$ the $(i,j)$-th component of the matrix $K$. Finding such an optimal subset $L$ is infeasible in practice, and the ProtoDash Explainer algorithm helps us to find an approximate solution heuristically. It starts by setting $L = \emptyset$ and $w_j=0 \ \forall j \in I$. Each iteration of the ProtoDash Explainer consists of two steps. In the first step, the index $j_0 \not \in L$ that takes the maximum value in $g =\nabla l(w) = \mu - K w$ is selected. In the second step,  the set of weights $w$ is updated to maximize $l(w)$, subject to $w_j \geq 0 \ \forall j \in I$ and $w_j=0 \ \forall j\not \in L$. The algorithm ends when $|L|=m$ and the output subset is $B=\{a_j \in A : j \in L\}$. The ProtoDash Explainer does not necessarily give an optimum solution, but it is proven in \cite{gurumoorthy2019efficient} that the quality of the approximate solution is lower-bounded by a fraction of the quality of the optimum solution. The pseudocode for the ProtoDash Explainer can be seen in Algorithm \ref{alg:protodash}.

Then, the ProtoDash Selection algorithm just applies the ProtoDash Explainer to each class $X_k$, finding a subset $S_k \subset X_k$ with $n_k$ examples that approximately minimizes $\widehat{MMD}(K,X_k,S_k,w)$. The reduced dataset $\dd_R$ is the union of all $S_k$. The pseudocode for Protodash Selection can be seen in Algorithm \ref{alg:PRD}.

\begin{algorithm}[t]
\caption{ProtoDash explainer}
\label{alg:protodash}
\KwData{$A = \{a_i\}_{i=1}^{n_A} \subset \erre^d$, \ $K:A \times A \rightarrow \erre$, \ $1\leq m \leq n_A$}
\KwResult{$B \subset A$}
Set $I = \{1,\cdots,n_A\}$ the set of indices in $A$\;
Set $L = \emptyset$ the set of selected indices\;
Set $\text{K}_{i,j}=K(a_i,a_j) \  \forall i,j \in I$\;
Set $\mu_{j}=\frac{1}{n_A}\sum_{i}K(a_i,a_j)  \ \forall j\in I$\;
Set $w_j=0 \  \forall j\in I$\;
Define $l(w) = w^T  \mu - \frac{1}{2} w^T  \text{K}  w$\;
Define $\nabla l(w)=\mu - \text{K} w$\;
\While{$|L|<m$}{
Set $g =\nabla l(w)$\;
Set $j_0 = \arg \max_{j\in I \setminus L} g_j$\;
Update $L = L \cup \{j_0\}$\;
Solve $\xi = \arg \max_{w} l(w)$ subject to $w_j\geq 0  \ \forall j\in I$, $w_j=0 \  \forall j\not \in L$\;
Update $w = \xi$\;
}
Set $B=\{a_j \in A : j \in L\}$\;
\end{algorithm}

\begin{algorithm}
\caption{PRD: ProtoDash Selection}
\label{alg:PRD}
\KwData{$\dd=(X,f)$, \ $p\in [0,1]$, \ $K:X \times X \rightarrow \erre$}
\KwResult{$\dd_R=(S,g)$}
\For{$k=1,\cdots,c$}{
Set the number of examples to be selected as $n_k = \lfloor p \cdot |X_k| \rfloor$\;
Apply Algorithm \ref{alg:protodash} with $A = X_k$ and $m = n_k$ to get $S_k = B$
}
Set $S=\bigcup_{k=1,\cdots,c}S_k$\;
Set $g = f|_{S}$\;
\end{algorithm}

\subsection{Geometry-based Methods}

In this subsection, we introduce three data reduction methods that use the distances between the examples in $\dd$ to find a reduced dataset $\dd_R$. 

\subsubsection*{Clustering Centroids Selection (CLC)}

Clustering is a branch of unsupervised machine learning whose task is to partition a dataset into groups or clusters, where objects within the same cluster are highly similar and distinct from those in other clusters. The goal is to discover patterns or structures without prior knowledge or labels. Clustering algorithms yield diverse partitions depending on the approach. For a comprehensive overview of clustering, we refer to \cite{xu2008clustering}.

The Clustering Centroids Selection (CLC) algorithm proposes to use $k$-means, one of the most-known clustering algorithms, for data reduction. This idea was stated in 
\cite{bezdek2001nearest} and \cite{liu2002issues}, among others.
The general idea is to apply $k$-means on each class of $\dd$ and include the resulting centroids in $\dd_R$. This is the only data reduction method in
this paper where the reduced dataset $\dd_R$ is not necessarily a sub-dataset of $\dd$. This method is easy to understand but can be computationally expensive for large datasets, unstable, and sensitive to outliers, as stated in \cite{chawla2013k, li2011k}. The pseudocode for CLC can be read in Algorithm \ref{alg:CLC}.

\begin{algorithm}
\caption{CLC: Clustering Centroids Selection}
\label{alg:CLC}
\KwData{$\dd=(X,f)$, \ $p\in [0,1]$}
\KwResult{$\dd_R=(S,g)$}
\For{$k=1,\cdots,c$}{
Set the class $k$ as $X_k=\{x \in X: f(x)=k\}$\;
Set the number of examples to be selected as $n_k = \lfloor p \cdot |X_k| \rfloor$\;
Apply $k$-Means on $X_k$ with $n_k$ clusters\;
Select the set of centroids $S_k$\;
}
Set $S=\bigcup_{k=1,\cdots,c}S_k$\;
\For{$k=1,\cdots,c$}{
Set $g(x)=k$ for each $x \in S_k$\;
}
\end{algorithm}

\subsubsection*{Maxmin Selection (MMS)}

Maxmin Selection (MMS) uses the distances between the examples to ensure that $\dd_R$ is evenly spaced. It has been used in \cite{lacombe_data-driven_2021} for 
the reduction of datasets
and in \cite{de2004topological} to create efficient data descriptors. For each class $k$, the first step is to pick a random example $x_r \in X_k$ and add it to $S_k$. Then, given a distance function $d:X \times X \rightarrow \erre^+$, each step picks the example in $X_k \setminus S_k$ that maximizes the function:

\begin{align*}
    D:X_k \setminus S_k & \rightarrow \erre \\
    x & \mapsto \min_{x' \in S_k} d(x,x')
\end{align*}

This is repeated until $S_k$ has the required size. This method gives a subset that covers up well the dataset, but tends to pick extreme or outlier points. The pseudocode for MMS can be read in Algorithm \ref{alg:MMS}.

\begin{algorithm}
\caption{MMS: Maxmin Selection}
\label{alg:MMS}
\KwData{$\dd=(X,f)$, \ $p\in [0,1]$, \ $d:X \times X \rightarrow \erre^+$}
\KwResult{$\dd_R=(S,g)$}
\For{$k=1,\cdots,c$}{
Set the class $k$ as $X_k=\{x \in X: f(x)=k\}$\;
Set the number of examples to be selected as $n_k = \lfloor p \cdot |X_k| \rfloor$\;
Select a random example $x_r \in X_k$\;
Set $S_k = \{x_r\}$\;
\While{$|S_k|<n_k$}{
Set $x = \arg \max_{x \in X_k \setminus S_k} \min_{x' \in S_k} d(x,x')$\;
Update $S_k = S_k \cup \{x\}$\;
}
}
Set $S=\bigcup_{k=1,\cdots,c}S_k$\;
Set $g = f|_{S}$\;
\end{algorithm}

\subsubsection*{Distance-Entropy Selection (DES)}

Distance-Entropy Selection (DES) \cite{li_distance-entropy_2022} is a data reduction method that tries to ensure that the resulting dataset $\dd_R$ has relevant examples. It is based on a distance-entropy indicator that measures how informative are the different examples for the classification task. 

The algorithm begins by selecting a subset $X_{base} \subset X$, known as the base data. In our implementation, we have decided to select the base data via Stratified Random Selection, using a proportion $p_{base} < p$. The base data is used to calculate a prototype $p_k$ for each class $k=1,\cdots,c$. In our case, the prototype $p_k$ is defined as the average of all points of class $k$ in $X_{base}$. The algorithm then calculates the distances between the prototypes and all points in $ X_{pool} = X \setminus X_{base}$, called the pool data. The distances $d_k=d(x,p_k)$ associated to $x \in X_{pool}$ are transformed into a probability distribution by the softmax function, with formula:

\begin{equation}
\operatorname{Softmax}(d_k) = \frac{e^{d_k}}{\sum_{j=1}^c e^{d_j}}
\end{equation}

The information entropy of this distribution is called the distance-entropy indicator of $x$: 

\begin{equation}
E(x) = - \sum_{k=1}^c \operatorname{Softmax}(d_k) \cdot \log_2 \operatorname{Softmax}(d_k)
\end{equation}

Finally, the reduced dataset $\dd_R$ is formed by all the examples in $X_{base}$ and the examples from $X_{pool}$ with the highest values for the distance-entropy indicator. The pseudocode for Distance-Entropy Selection can be seen in Algorithm \ref{alg:DES}.

\begin{algorithm}
\caption{DES: Distance-Entropy Selection}
\label{alg:DES}
\KwData{$\dd=(X,f)$, \ $p\in [0,1]$,\ $p_{base} \in [0,p]$ ,\ $d:X \times X \rightarrow \erre^+$}
\KwResult{$\dd_R=(S,g)$}
\For{$k=1,\cdots,c$}{
Set the class $k$ as $X_k=\{x \in X: f(x)=k\}$\;
Set the number of examples to be included in the base data as $n_{k,base} = \lfloor p_{base} \cdot |X_k| \rfloor$\;
Select a random subset $S_{k,base} \subset X_k$ with $|S_k|=n_{k,base}$ \;
Calculate a prototype $p_k$ for the examples in $S_{k,base}$\;
}
Set $X_{base} = \bigcup_{k=1,\cdots,c} S_{k,base}$\;
Set $X_{pool} = X \setminus X_{base}$\;
\For{$x \in X_{pool}$}{
\For{$k=1,\cdots,c$}{
Calculate the distance $d_k=d(x,p_k)$\;
}
\For{$k=1,\cdots,c$}{
Transform the distances into probabilities with $\operatorname{Softmax}(d_k) = \frac{e^{d_k}}{\sum_{j=1}^c e^{d_j}}$\;
}
Calculate the distance-entropy indicator as $E(x) = - \sum_{k=1}^c \operatorname{Softmax}(d_k) \cdot \log_2 \operatorname{Softmax}(d_k)$\;
}
Set the number of examples to be added as $n_{add} = \lfloor p \cdot |X| \rfloor - |X_{base}|$\;
Set $X_{add} \subset X_{pool}$ containing the $n_{add}$ examples in $X_{pool}$ with higher values for $E$\;
Set $S=X_{base} \cup X_{add}$\;
Set $g = f|_{S}$\;
\end{algorithm}

To justify why the best examples are those with higher entropies, the authors of \cite{li_distance-entropy_2022} use the following reasoning. Suppose that an example $x \in X_{pool}$ is closer to one prototype $p_k$ than to all the others. In that case, the distance-entropy indicator $E(x)$ will be low and $x$ is likely to be classified in class $k$. By contrast, items with high entropy are informative because they are different from all prototypes and not so easy to classify. 
Note that the examples from the pool data are selected regardless of their class, so it is possible that the reduction ratio of each class is different from the global reduction ratio. For this reason, we recommend selecting the base data using a sufficiently high $p_{base}$ to make sure that all classes are well represented and then complementing the base data with the most informative examples from the pool data.

\subsection{Ranking-based Methods}

In this subsection, we describe three methods that are based on a ranking system. Basically, these methods assign a score to the examples based on a particular criterion, sort them according to their score, and then select the best-ranked examples from this sorted list.

\subsubsection*{PH Landmarks Selection (PHL)}

PH Landmarks Selection (PHL) \cite{stolz2023outlier} is a subset selection method based on the concept of persistent homology. Roughly speaking, persistent homology is a common technique in topological data analysis (TDA) that builds a filtration of simplicial complexes over the dataset examples (such as Vietoris-Rips filtration) and computes for each $n \geq 0$ the evolution of certain mathematical features (called $n$-dimensional homology classes) along the filtration. The $n$-dimensional persistent homology of a data set can be encoded in a barcode $B_n = \{[b_i,d_i)\}_{i=1}^{I_n}$ that has a bar $[b,d)$ for each $n$-dimensional homology class that first appears in the stage $b$ of the filtration and disappears at stage $d$.

PHL algorithm orders the examples in each class by evaluating how their removal changes its persistent homology. Given an example $x \in X_k$, the first step is to find its $\delta$-neighbourhood \(\Delta_x=\{\xtilde \in X_k \setminus \{x\}: d(x,\xtilde)\leq \delta  \}\). If \(|\Delta_x| \leq 2\), $x$ is considered a super-outlier. If $x$ is not a super-outlier, a Vietoris-Rips filtration is built over $\Delta_x$ and its persistent homology is computed for $n=0,1,2$. Then, the \textit{PH outlierness} of $x$ is: 
\begin{equation}
out_{PH}^{0,1,2}(x) = \max_{n=0,1,2} \max_{i}\{d_i-b_i: [b_i,d_i) \in B_n(\Delta_x)\}
\end{equation}
A restricted version of PH outlierness that can be used in practice is:
\begin{equation}
out_{PH}^{1}(x) = \max_{i}\{d_i-b_i: [b_i,d_i) \in B_1(\Delta_x)\}
\end{equation}

We denote the PH outlierness as $out_{PH}$. Small values for $out_{PH}(x)$ indicate that the persistent homologies of $X_k$ and $X_k \setminus \{x\}$ are similar. The theoretical motivation for this statement can be found in \cite{stolz2023outlier}. 
Two strategies are proposed to select examples from $X_k$. On the one hand, we can choose the examples that are not super-outliers and have smaller values for $out_{PH}(x)$, called representative landmarks. On the other hand, we can choose those with higher values for $out_{PH}(x)$, called vital landmarks. In case there are not enough examples in $X_k$ that are not super-outliers to be chosen, the subset can include some random super-outliers. The reduced dataset $\dd_R$ is created by applying this procedure for each class. 
Algorithm~\ref{alg:PHL} shows the pseudocode for PHL selection.

\begin{algorithm}
\caption{PHL: PH Landmarks Selection}
\label{alg:PHL}
\KwData{$\dd=(X,f)$, \ $p\in [0,1]$,\ $d:X \times X \rightarrow \erre^+$, \ $\delta > 0$, $o_{type} \in \{\text{
multidimensional}, \text{restricted}\}$,\ $l_{type} \in \{\text{representative},\text{vital}\}$}
\KwResult{$\dd_R=(S,g)$}
\For{$k=1,\cdots,c$}{
Set the class $k$ as $X_k=\{x \in X: f(x)=k\}$\;
Set the number of examples to be selected as $n_{k} = \lfloor p \cdot |X_k| \rfloor$\;
Set $O_k = \emptyset$ the set of super-outliers\;
\eIf{$o_{type} = \text{multidimensional}$}{
Set $out_{PH} \equiv out_{PH}^{0,1,2}$\;
}{
Set $out_{PH} \equiv out_{PH}^{1}$\;
}
\For{$x \in X_k$}{
Find $\Delta_x = \{\xtilde \in X_k \setminus \{x\}: d(x,\xtilde)\leq \delta  \}$\;
\eIf{$|\Delta_x|>2$}{
Compute the Vietoris-Rips filtration of $\Delta_x$ for $n=0,1,2$\;
Compute $out_{PH}(x)$\;
}{
Update $O_k = O_k \cup \{x\}$\;
}
}
\eIf{$n_k \leq |X_k \setminus O_k|$}{
\eIf{$l_{type} = \text{representative}$}{
Set the subset $S_k \subset X_k \setminus O_k$  with the $n_k$ lowest values for $out_{PH}$\;
}{
Set the subset $S_k \subset X_k \setminus O_k$  with the $n_k$ highest values for $out_{PH}$\;
}
}{
Select a random subset $R_k \subset O_k$ with $|R_k|=|X_k| - n_k$ \;
Set $S_k = X_k \setminus R_k$\;
}
}
Set $S=\bigcup_{k=1,\cdots,c}S_k$\;
Set $g = f|_{S}$\;
\end{algorithm}

\subsubsection*{Numerosity Reduction by Matrix Decomposition (NRMD)}

\begin{algorithm}
\caption{Calculate scores from a matrix}
\label{alg:scores}
\KwData{$A \in \erre^{N \times d}$, $d_{type} \in \{\text{SVD},\text{NMF},\text{PLU},\text{QR},\text{DL},\text{SPCA},\text{FLDA}\}$}
\KwResult{$s \in \erre^N$}
Set $r = \min \{N,d\}$\;
Calculate $A = UV$ using the $d_{type}$ decomposition\;
Set $\Tilde{A}$ as the normalization of $A$\;
Set $\Tilde{V}$ as the normalization of $V$\;
\eIf{$d_{type} \in \{\text{SVD},\text{SPCA},\text{FLDA}\}$}{
Set $w \in \erre^{r}$ with $w_i=\frac{1/\lambda_i}{\sum_i 1/\lambda_i}$, being $\lambda_1> \cdots> \lambda_r$ the eigenvalues given by the decomposition\;
}{
Set $w \in \erre^{r}$ with $w_i=\frac{2i}{r(r+1)}$\;
}
Calculate the scores vector $s = - \log(|\Tilde{A}\Tilde{V}^T|_{\varepsilon})w$\;
\end{algorithm}

Numerosity Reduction by Matrix Decomposition (NRMD) \cite{ghojogh2019instance} is a method that leverages matrix decomposition to rank examples in a dataset $\dd=(X,f)$. To use this method, it is necessary to use the tabular representation of $\dd$ that we saw in Subsection
\ref{subsec:tabulardata}. The matrix $\mathbf{X}$ contains all the examples in $X$, and the submatrix $\mathbf{X}_k$
contains all the examples in $X_k$.

Given a matrix $A\in \erre^{n\times d}$ with rows $a_1,\cdots,a_n$, a decomposition is just a factorization $A=UV$, where $U\in \erre^{n\times r}$, $V\in \erre^{r\times d}$ with rows $v_1,\cdots,v_r$, and $r=\min\{n,d\}$. Some typical matrix decompositions are Singular Value Decomposition (SVD) \cite{golub1971singular}, Non-negative Matrix Factorization (NMF) \cite{lee2000algorithms}, PLU Decomposition \cite{golub2013matrix}, QR decomposition \cite{golub2013matrix}, Dictionary Learning (DICL) \cite{mairal2009online}, Supervised Principal Component Analysis (SPCA) \cite{barshan2011supervised} and Fisher Linear Discriminant Analysis (FLDA) \cite{xanthopoulos2013linear}. From this decomposition, each row of $A$ is assigned a score based on its similarity to the rows of $V$. The matrix $\Sigma$, with $\Sigma_{i,j}= |\cos(a_i,v_j)|_{\varepsilon}$, stores all the similarities ($|\cdot|_{\varepsilon}$ denotes 
the maximum between the absolute value and a certain $\varepsilon>0$). The final score vector is $-\log(\Sigma)w$, where $w \in \erre^r$ is a weight vector given by $w_i=\frac{1/\lambda_i}{\sum_i 1/\lambda_i}$ when the decomposition is based on eigenvalues (as in SVD, SPCA and FLDA) and by $w_i=\frac{2i}{r(r+1)}$ otherwise (as in NMF, DL, PLU and QR decompositions). 
Algorithm \ref{alg:scores} shows the
procedure to calculate scores from a matrix decomposition.

Given a specific decomposition type, the NRMD method computes scores for all matrices $\mathbf{X}_1, \cdots, \mathbf{X}_c$. As a result, there exists a score \(s_{\mathbf{X}}(x)\) for each \(x \in \mathbf{X}\). In addition, the method calculates scores \(s_{D}(x)\) for the matrix \(D = [\mathbf{X} | E]\), where \(E \in \mathbb{R}^{N\times c}\) represents the one-hot encoding matrix of \(f\). In this encoding, \(E_{ij} = 1\) if \(f(x_i) = j\) and \(E_{ij} = 0\) otherwise. The final score for an example $x\in X$ is \( s(x) = s_{\mathbf{X}}(x) \cdot s_{D}(x)\). The dataset
$\dd_R$ is finally formed by the examples with the highest values for $s$, which are considered the most useful in terms of internal representation and discrimination between classes. The pseudocode for Numerosity Reduction by Matrix Decomposition can be seen in Algorithm \ref{alg:NRMD}.

\begin{algorithm}
\caption{NRMD: Numerosity Reduction by Matrix Decomposition}
\label{alg:NRMD}
\KwData{$\dd=(X,f)$, \ $p\in [0,1]$, \ $d_{type} \in \{\text{SVD},\text{NMF},\text{PLU},\text{QR},\text{DICL},\text{SPCA},\text{FLDA}\}$}
\KwResult{$\dd_R=(S,g)$}
Set $n = \lfloor p \cdot |X|\rfloor$\;
\For{$k=1,\cdots,c$}{
Set the class $k$ as $X_k=\{x \in X: f(x)=k\}$\;
Calculate $s_{X_k}$ as the result of applying Algorithm \ref{alg:scores} to the matrix $\mathbf{X}_k$\;
}
Obtain $s_{\mathbf{X}}$ merging the score vectors $s_{X_1},\cdots,s_{X_c}$\;
Calculate \(E\) as the one-hot encoding matrix of \(f\)\;
Set $D = [\mathbf{X}|E]$\;
Calculate $s_{D}$ as the result of applying Algorithm \ref{alg:scores} to the matrix $D$\;
Calculate the final scores $s = s_{\mathbf{X}} \odot s_D$\;
Set the subset $S \subset X$ with the $n$ highest scores in $s$\;
Set $g = f|_{S}$\;
\end{algorithm}

\subsection{Wrapper Methods}

All the data reduction methods described in the previous subsections are intended to be applied before training $\mathcal{N}$, since they only need the information from $\dd$ itself to extract $\dd_R$. In this section, we describe a method that uses the information obtained during training of $\mathcal{N}$ to reduce $\dd$. This means that the data reduction is not done before the training, but is wrapped in the training process itself.

\subsubsection*{Forgetting Events Selection (FES)}

Forgetting Events Selection (FES) is a data reduction method that leverages the evolution of accuracy throughout neural network training. During the training process, an example $x \in X$ can be well classified after some epochs (we say that its current accuracy is $a_x=1$) and misclassified after others (we say that $a_x=0$). If $a_x=0$ after epoch $t-1$ but $a_x=1$ after epoch $t$, we say that $x$ has undergone a \textit{learning event}. Conversely, if $a_x=1$ after epoch $t-1$ but $a_x=0$ after epoch $t$, $x$ has undergone a \textit{forgetting event}. \textit{Unforgettable examples} are those with $a_x=1$ that never had a forgetting event. 

The experiments in \cite{toneva2018empirical} show that unforgettable examples have less impact on network training than those that go through several forgetting events and that they can be removed from the training dataset without significantly affecting the model performance. Based on this idea, the FES algorithm counts how many forgetting events each example undergoes during training and discards the examples with the lowest number of forgetting events. Examples that never get well classified are assigned an infinite number of forgetting events.

Following the ideas from \cite{coleman2019selection}, our FES implementation only counts the forgetting events during the first $e_{initial}$ epochs of the training process. At that point, the algorithm reduces $\dd$ by selecting examples with more forgetting events and performs the remaining epochs using only $\dd_R$ as a training dataset. To ensure that all classes are well represented in $\dd_R$, the selection is made class by class. Algorithm \ref{alg:FES} shows how to apply FES selection during the training of a DL model $\mathcal{N}$.

\begin{algorithm}[H]
\caption{FES: Forgetting Events Selection}
\label{alg:FES}
\KwData{$\dd=(X,f)$, \ $p\in [0,1]$, \ $\mathcal{N}:\erre^d \rightarrow \{1,\cdots,c\}$, $e_{initial}$, $e_{total}$}
\KwResult{$\dd_R=(S,g)$, $\mathcal{N}$}
\For{$x \in X$}{
Set the current accuracy $a_x = 0$\;
Set the number of forgetting events $f_x = 0$\;
}
\For{$e=1,\cdots,e_{initial}$}{
Perform a training epoch on $\mathcal{N}$ using $\dd$\;
\For{$x \in X$}{
\If{$\mathcal{N}(x)=f(x)$}{
Update $a_x=1$\;
}
\ElseIf{$a_x=1$}{
Update $f_x = f_x + 1$\;
Update $a_x = 0$\;
}
}
}
\For{$x \in X$}{
\If{$a_x=f_x=0$}{
Update $f_x = \infty$
}
}
\For{$k=1,\cdots,c$}{
Set the class $k$ as $X_k=\{x \in X: f(x)=k\}$\;
Set the number of examples to be selected as $n_{k} = \lfloor p \cdot |X_k| \rfloor$\;
Select a subset $S_k \subset X_k$ with the $n_k$ highest values for $f_x$\;
}
Set $S=\bigcup_{k=1,\cdots,c}S_k$\;
Set $g = f|_{S}$\;
\For{$e=e_{initial}+1,\cdots,e_{total}$}{
Perform a training epoch on $\mathcal{N}$ using $\dd_R = (S,g)$\;
}
\end{algorithm}
\section{Experiments}\label{sec:experiments}

In this section, we present the datasets used for the experiments,
the parameter settings, the setup for the experiments, and finally,
the obtained results. The source code of the experiments is available in the GitHub repository \cite{Victor_Toscano_Duran_Repository_experiments_Survey_2023}.

\subsection{Experiments for Tabular Data Classification}

In this subsection, we describe the two experiments that we have developed to analyze the utility of data reduction for classification tasks with tabular datasets. In the first place, we detail the methodology to apply the different data reduction methods to a dataset and measure its efficiency, representativeness and performance. Then, we give some details on the two datasets we used. Finally, we show the results obtained for both experiments and discuss the main conclusions.

\subsubsection{Datasets for Classification}

The two datasets that we have used in our experiments are:

\paragraph*{Collision dataset} 

This tabular dataset was provided by our colleagues Maurizio Mongelli and Sara Narteni, based on \cite{collision}. It can be downloaded from the repository \cite{Victor_Toscano_Duran_Repository_experiments_Survey_2023}, that also contains the code for the experiments and our results.
The classification task consists of predicting whether a platoon of vehicles will collide based on features such as the number of cars and their speed. The dataset consists of 107,210 examples with 25 numerical features and 2 classes:
\begin{itemize}
\item collision = 1, with 69,348 examples.
\item collision = 0, with 37,862 examples.
\end{itemize}

We decided to use this dataset in experiments to test the usefulness of data reduction methods to reduce resource consumption in a task related to safe mobility. Before the experiments, we discarded the two features ``N'' and ``m'' since they are constant and do not help us in the classification task. 

\paragraph*{Dry Bean dataset} This dataset (see \cite{misc_dry_bean_dataset_602} and \cite{KOKLU2020105507}) was created by taking pictures of dry beans from 7 different types and calculating some geometric features from the images, such as the area, the perimeter and the eccentricity. The classification task consists of predicting the type of dry bean based on these geometric features. The dataset contains 13,611 examples with 16 features and 7 classes:

\begin{itemize}
\item Barbunya, with 1,322 examples.
\item Bombay, with 522 examples.
\item Cali, with 1,630 examples.
\item Dermason, with 3,546 examples.
\item Horoz, with 1,928 examples.
\item Seker, with 2,027 examples.
\item Sira, with 2,636 examples.
\end{itemize}
 
The classes were encoded from 0 to 6 for the experimentation following the listed ordering. We decided to use this dataset to test the usefulness of data reduction methods for classification tasks with several unbalanced classes.

\subsubsection{Methodology}

The methodology of the experiments for both datasets consists of the following
three 
steps:

\begin{enumerate}
\item \textbf{Dataset Preprocessing:} 

It is a common practice to scale or standardize a dataset before building the DL model because this increases the likelihood that the training process will be fast and will not be conditioned by some features simply due to their greater magnitude \cite{ahsan2021effect, sharma2022study}. In our case, we decided to apply the scikit-learn function \textit{MinMaxScaler} \cite{scikit-learn}.

Each feature $p_j$ has a maximum value $p_{j,max} = \max_{x_i \in X} p_j(x_i)$ and a minimum value $p_{j,min} = \min_{x_i \in X}  \newline p_j(x_i)$. We say that the range of $p_j$ is the interval $\left[p_{j,min},p_{j,max}\right]$ because all its possible values lie in it. Rescaling with \textit{MinMaxScaler} is just changing the example $x_i \in X$ by:

\begin{equation}
    x_{i,scaled} = \left( \frac{x_{i,1}-p_{1,min}}{p_{1,max}-p_{1,min}}, \cdots, \frac{x_{i,j}-p_{j,min}}{p_{j,max}-p_{j,min}}, \cdots, \frac{x_{i,d}-p_{d,min}}{p_{d,max}-p_{d,min}} \right)^T
\end{equation}

After scaling, the range of all the features is $[0,1]$. That means that all of them have similar values and can be compared between them. 

\item \textbf{Fixing the architecture and hyperparameters:} 

In both experiments, we used a neural architecture with 10 layers with the following dimensions: 

\begin{equation}
X \xrightarrow{f_1} \mathbb{R}^{50} \xrightarrow{f_2} \mathbb{R}^{45} \xrightarrow{f_3} \mathbb{R}^{40} \xrightarrow{f_4} \mathbb{R}^{35} \xrightarrow{f_5} \mathbb{R}^{30} \xrightarrow{f_6} \mathbb{R}^{25} \xrightarrow{f_7} \mathbb{R}^{20} \xrightarrow{f_8} \mathbb{R}^{15} \xrightarrow{f_9} \mathbb{R}^{10} \xrightarrow{f_{10}} O
\end{equation}

All layers except the last one, use the Rectified Linear Unit (ReLU) activation function $\text{ReLU}(x) = \max(0, x)$. Additionally, these layers use dropout as a regularization technique. 
The probability of zeroing a neuron during dropout each time is a hyperparameter called the \textit{dropout probability}, which we have set equal to 0.50 for the experiment with the Collision dataset and equal to 0.25 for the experiment with the Dry Bean dataset. The differences in neural architecture for both experiments are in the last layer and in the output space $O$.

For the Collision dataset, the last layer has only one neuron (that is, $O=\erre$), with sigmoid activation function $\sigma(x) = 1/(1 + e^{-x})$. The output of $\mathcal{N}$ (also called the logit) for an input $x_i$ is a number $z_i \in [0,1]$. The predicted class for $x_i$ is:

\begin{equation}
\mathcal{N}_{\theta}(x_i)= \left\{
\begin{array}{ll}
    0 & \text{if} \ z_i < 0.5 \\
    1 &  \text{otherwise}
\end{array}
\right.
\end{equation}

To train the network for the Collision dataset, we used the Binary Cross Entropy as a loss function. Given a neural network $\mathcal{N}_{\theta}$ and a dataset $\dd$, it has the following formula:

\begin{equation}
\text{BCELoss}(\theta,\dd) = -\frac{1}{N} \sum_{i=1}^{N} \left(f(x_i) \cdot \log(z_i) + (1 - f(x_i)) \cdot \log(1 - z_i) \right)
\end{equation}

For the Dry Bean dataset, the last layer has 7 neurons, one for each class (that is, $O=\erre^7$). The logits $z_{i,0},\cdots,z_{i,6}$ for an input $x_i$ are transformed into a probability distribution with the softmax activation function, given by $s_{i,k} = \text{softmax}(z_{i,k}) = e^{z_{i,k}}/\sum_{m=0}^{6} e^{z_{i,m}}$, and the predicted class for $x_i$ is:

\begin{equation}
\mathcal{N}_{\theta}(x_i)= \arg \max_{k=0,\cdots,6} s_{i,k}
\end{equation}

To train the network for the Dry Bean dataset, we used the Categorical Cross Entropy loss function. If we denote $y_{i,k} = 1$ if $f(x_i)=k$ and $y_{i,k} = 0$ otherwise, the formula of Categorical Cross Entropy is:

\begin{equation}
\text{CCELoss}(\theta,\dd) = -\frac{1}{N} \sum_{i=1}^{N} \sum_{k=0}^{6} w_k \cdot y_{i,k} \cdot \log(s_{i,k})
\end{equation}

Here, $w_k = N/N_k$ is a weight assigned to class $k$ to give more importance to the least populated classes and prevent a bias towards the most populated ones. 

In both cases, we used the Adam optimizer \cite{Kingma2014AdamAM} to minimize the loss function, specifying a learning rate of $\gamma = 0.001$ and letting the default values for the rest of the required hyperparameters.

Regarding the other learning hyperparameters, the network for the Collision dataset was trained for $n_e = 600$ epochs with a batch size of $\beta = 1,024$. When the FES reduction was applied, the model was trained for $n_i = 200$ epochs with the full training dataset and the remaining $400$ epochs with the reduced dataset. For the Dry Bean dataset, the number of training epochs was $n_e=150$ ($n_i=50$ for the first part of training with FES reduction) and the batch size was of $\beta=32$.

\item \textbf{Data reduction and model training:} Now that the dataset is scaled and the neural architecture and the learning hyperparameters have been set, in this step we analyze how the data reduction methods affect the efficiency and performance of the training of a neural network. This step is divided into the following 4 sub-steps:

\begin{itemize}
\item \textbf{Train-Test dataset split:} The dataset $\dd$ is randomly split into a training dataset $\dd_{train}$ and a test dataset $\dd_{test}$. The DL model will be trained using $\dd_{train}$ and its performance will be evaluated using $\dd_{test}$. The test dataset contains a proportion $p_{test} \in (0,1)$ of the total number of examples in $\dd$. For both experiments, we set $p_{test}=0.25$.
\item \textbf{Training with no reduction:} In this step the DL model is trained using the whole training dataset $\dd_{train}$ with no reduction, and then, the computation time and carbon emission of the training are calculated.
After that, the model is used to classify the test dataset $\dd_{train}$ and the accuracy, macro average precision, macro average recall and macro average F1-score are computed.
\item \textbf{Training + reduction for non-wrapper methods:} In this step, $\dd_{train}$ is reduced getting $\dd_{train,R}$ as a result, and the $\varepsilon$-representativeness of $\dd_{train,R}$ respect to $\dd_{train}$ is computed. The model is then trained for $n_e$ epochs using $\dd_{train,R}$, and the total computation time and carbon emission of the reduction and the training are calculated.
The model is used to classify $\dd_{test}$ as in the previous step. This is repeated for each non-wrapper data reduction method (all but FES) and for each reduction percentage $p \in \{0.1,0.2,\cdots,0.9\}$.
\item \textbf{Training + reduction for FES:} In this step, the DL model is trained using $\dd_{train}$ for the first $n_i$ epochs. After applying the FES reduction, the model is trained for the remaining epochs using $\dd_{train,R}$, and the $\varepsilon$-representativeness of $\dd_{train,R}$ with respect to $\dd_{train}$ is also computed. 
The total computation time and carbon emission of the training and the reduction are computed. The model is used to classify $\dd_{test}$ as in the previous steps. 
This step is repeated for each reduction percentage $p \in \{0.1,0.2,\cdots,0.9\}$.
\end{itemize}
This last step is repeated 10 times to test how the data reduction works for different train-test splits and mitigate possible overfitting or bias caused by a specific split of the dataset.
\end{enumerate}

Algorithm \ref{algo:tab_pipeline} shows the experiment pipeline for tabular data classification.
\vspace{3mm}

\begin{algorithm}[H]
\KwData{$\dd=(X,f)$}
\textbf{Dataset Preprocessing}\;
Scale $\dd$ using MinMaxScaler\;
\textbf{Fixing the architecture and hyperparameters}\;
Set a test size proportion $p_{test} \in (0,1)$\;
Set a neural architecture and create the DL model $\mathcal{N}$\;
Set a loss function $\mathcal{L}:\Theta \rightarrow \erre^+$\;
Set an optimization algorithm to minimize $\mathcal{L}$ and its associated hyperparameters\;
Set a regularization technique and its associated hyperparameters\;
Set a number of training epochs $n_e \in \mathbb{N}$\;
Set a number of initial training epochs for FES reduction $n_i  \in \mathbb{N}$, with $n_i < n_e$\;
Set a batch size $\beta \in \mathbb{N}$\;
Set a number of iterations $n_{iter} \in \mathbb{N}$\;
\textbf{Data reduction and model training}\;
\For{$i=1$ \KwTo $n\_iter$}{
\textbf{Train-Test dataset split}\;
Set $N_{test} = \lfloor p_{test} \cdot N \rfloor$\;
Split $\dd$ into $\dd_{train}$ and $\dd_{test}$, being the size of $\dd_{test}$ equal to $N_{test}$\;
\textbf{Training with no reduction}\;
Train $\mathcal{N}$ for $n_e$ epochs using $\dd_{train}$\;
Calculate the computing time and carbon emission of the training\;
Validate the model with $\dd_{test}$ and calculate the accuracy, macro average precision, macro average recall and macro average F1-score\;
\textbf{Training + reduction for non-wrapper methods:}\;
\ForEach{non-wrapper method}{
\For{$p \in \{0.1,0.2,\cdots,0.9\}$}{
Get $\dd_{train,R}$ as the reduced dataset of $\dd_{train}$ with the corresponding data reduction method and the reduction ratio $p$\;
Calculate the $\varepsilon$-representativeness of $\dd_{train,R}$ respect to $\dd_{train}$\;
Train $\mathcal{N}$ for $n_e$ epochs using $\dd_{train,R}$\;
Calculate the computing time and carbon emission of the reduction and the training\;
Validate $\mathcal{N}$ using $\dd_{test}$ and calculate the accuracy, macro average precision, macro average recall and macro average F1-score\;
}
}
\textbf{Training + reduction for FES}\;
\For{$p \in \{0.1,0.2,\cdots,0.9\}$}{
Train $\mathcal{N}$ for $n_i$ epochs using $\dd_{train}$\;
Get $\dd_{train,R}$ as the reduced dataset of $\dd_{train}$ with FES reduction and the reduction ratio $p$\;
Calculate the $\varepsilon$-representativeness of $\dd_{train,R}$ respect to $\dd_{train}$\;
Train $\mathcal{N}$ for $n_e-n_i$ epochs using $\dd_{train,R}$\;
Calculate the computing time and carbon emission of the reduction and the training\;
Validate $\mathcal{N}$ using $\dd_{test}$ and calculate the accuracy, macro average precision, macro average recall and macro average F1-score\;
}
}
\caption{Pipeline of the Experiments for Tabular Data Classification}\label{algo:tab_pipeline}
\end{algorithm}

\subsubsection{Results and Discussion}

All results in this Section are the median values after $10$ repetitions. We chose to use the median for this experiment because it provides a robust measure of central tendency that is less affected by outliers, ensuring that our analysis is not influenced by extreme values.

\paragraph*{Collision Dataset}

\begin{figure}
        \centering
    \includegraphics[width = 0.7\textwidth]
{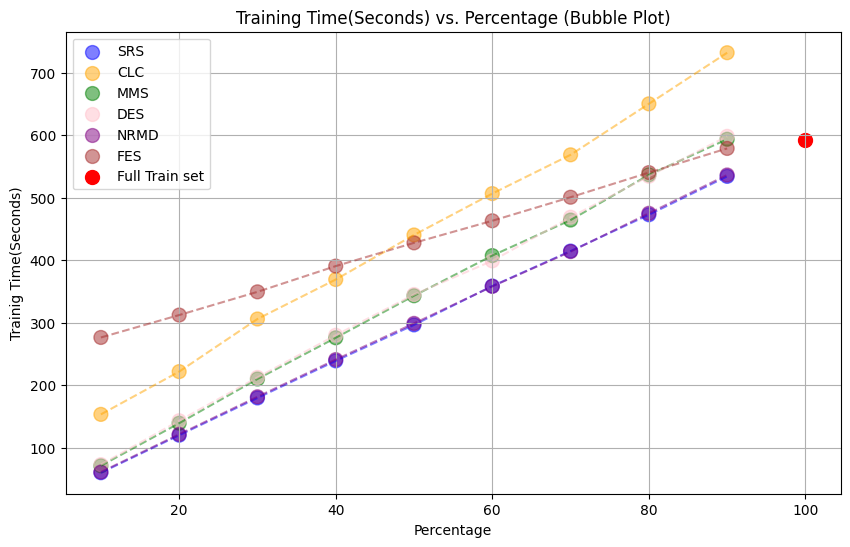}
        \caption[Collision: Reduction + Training time]{\label{collisionTT} Collision: Reduction + training time}
        \end{figure}
\begin{figure}
        \centering
    \includegraphics[width = 0.7\textwidth]
{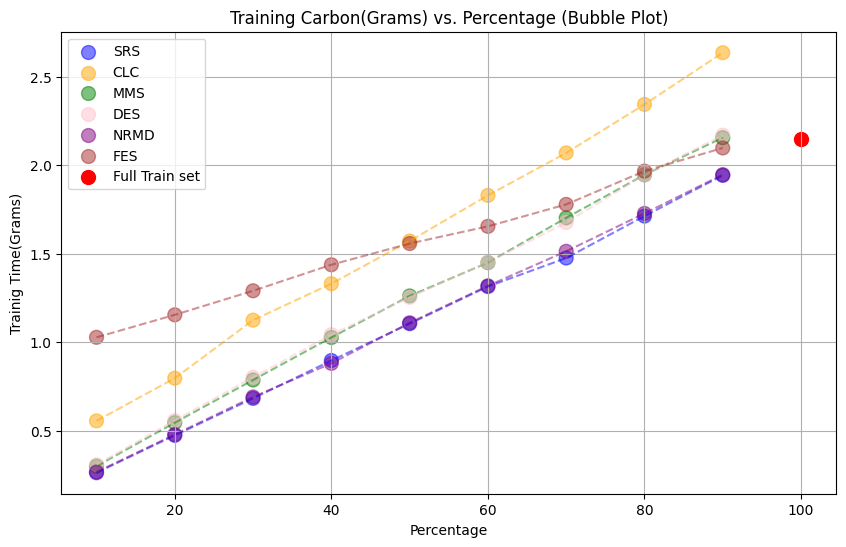}
        \caption[Collision: Reduction + Training carbon]{\label{collisionTC} Collision: Reduction + training carbon}
        \end{figure}

The median results we obtained for the efficiency metrics (computing time and carbon emission) can be seen in Figures \ref{collisionTT} and \ref{collisionTC}. The first thing that we can note is that both metrics express the same information since they are almost proportional. Approximately, each minute of computation during our experiment emitted $0.22$ g of $\text{CO}_2$ into the atmosphere. It is important to clarify that this occurs because we are primarily measuring CO2 emissions during the training time, which present the same characteristics regardless of the method used. However, as will be seen in the experiments for object detection, the CO2 emitted during the data reduction phase is not proportional to the computing time, because the reduction methods are not similar in terms of code. In general, the use of data reduction methods before network training helped to reduce the computing time and the carbon emission of model building with respect to the reference case (when the model is trained over the whole training dataset), but we find three particular exceptions. When the CLC reduction method is applied with a reduction ratio of $80\%$ or superior, the efficiency metrics are worse than those obtained for the reference case. We have the same situation for MMS and DES when the reduction ratio is equal to $90\%$. That suggests that, if we extract a reduced dataset with too many examples, it is possible that the time needed to compute the reduction does not compensate the time saved during the network training. Because of that, if the size of the dataset is equal to or larger than the size of the Collision dataset, we recommend applying the three reduction methods only to perform reductions with small reduction ratios. In all the other data reduction methods, we can observe that the efficiency metrics always improve those of the reference case. In the sense of efficiency, the two top data reduction methods are SRS and NRMD, with similar results in both the computation time and carbon emission. 
Additionally, we can see that the total time and carbon emission from the reduction and training are proportional to the number of examples in the reduced dataset. That indicates that the reduction takes a small time in the process, and almost all the measured time and carbon correspond to the network training. We observed that the efficiency of NRMD reduction depends on the type of matrix decomposition selected. We used SVD decomposition but the results may be different if we select another decomposition type.

\begin{figure}
        \centering
    \includegraphics[width = 0.7\textwidth]
{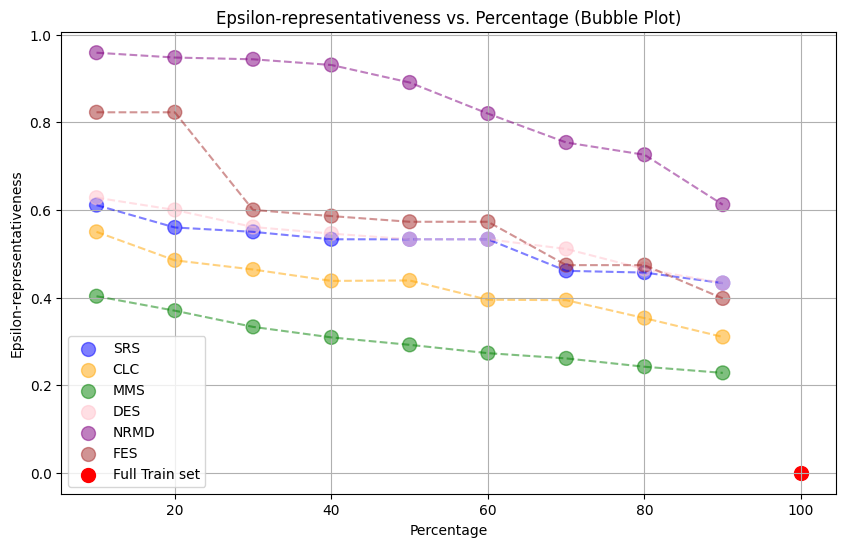}
        \caption[Collision: Reduction + $\varepsilon$-representativeness]{\label{collisionEpsilon} Collision: Reduction + $\varepsilon$-representativeness}
        \end{figure}

About the $\varepsilon$-representativeness of the reduced datasets with respect to the whole training dataset, the median 
results can be seen in Figure \ref{collisionEpsilon}.  The first thing we can observe is that MMS reduction is always the best at preserving the $\varepsilon$-representativeness for all the possible reduction ratios, which seems natural if we recall the definition of $\varepsilon$-representativeness and the way the MMS method selects each new example in the reduced dataset. CLC reduction also gives us datasets with good $\varepsilon$ values. In contrast, the data reduction method that has the highest $\varepsilon$ values for all the possible reduction ratios is NRMD.

\begin{figure}
        \centering
    \includegraphics[width = 0.7\textwidth]
{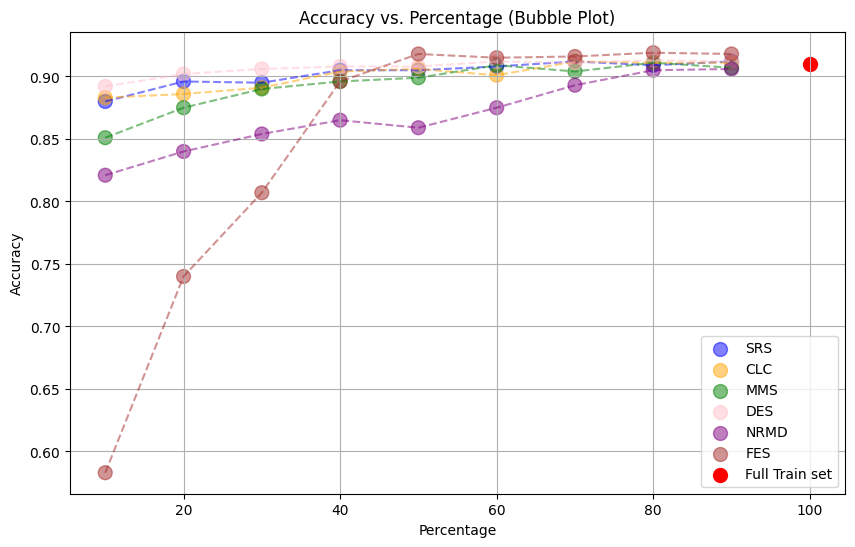}
        \caption[Collision: Reduction + Accuracy]{\label{collisionAccuracy} Collision: Reduction + Accuracy}
        \end{figure}

\begin{figure}
        \centering
    \includegraphics[width = 0.7\textwidth]
{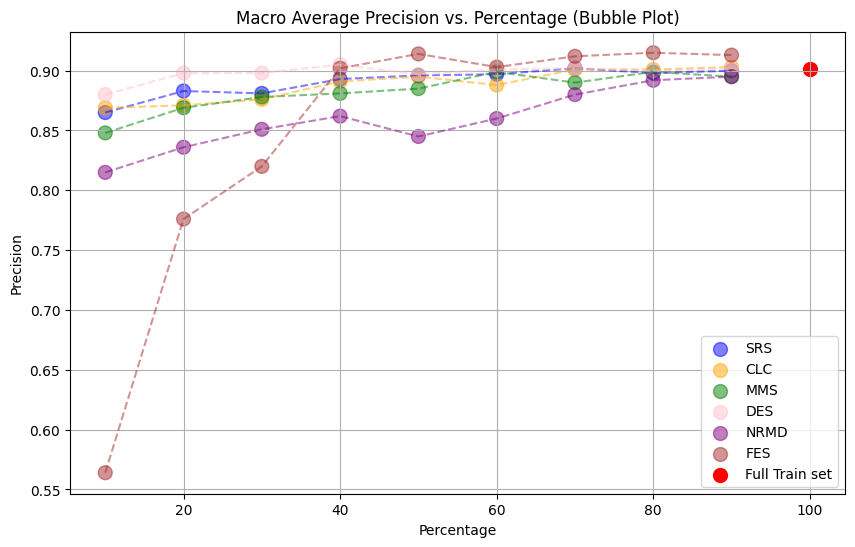}
        \caption[Collision: Reduction + macro average Precision]{\label{collisionPrecision} Collision: Reduction + macro average Precision}
        \end{figure}

\begin{figure}
        \centering
    \includegraphics[width = 0.7\textwidth]
{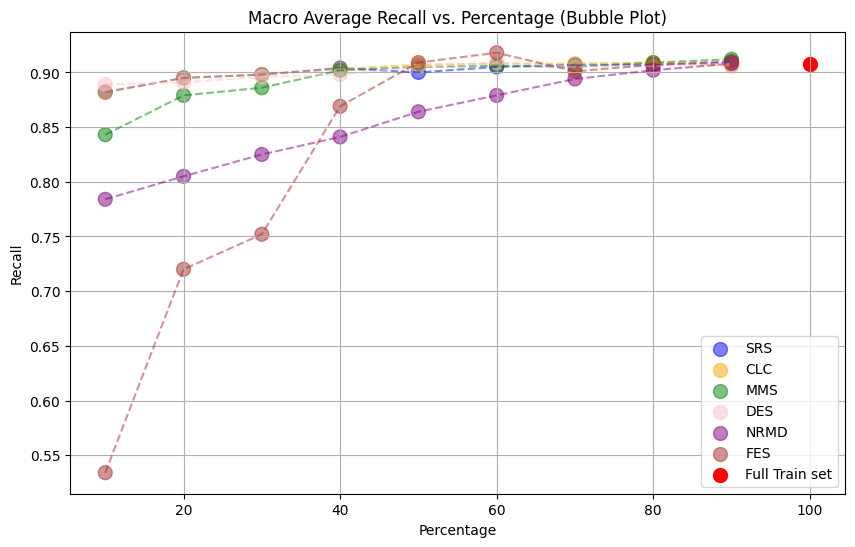}
        \caption[Collision: Reduction + macro average Recall]{\label{collisionRecall} Collision: Reduction + macro average Recall}
        \end{figure}

\begin{figure}
        \centering
    \includegraphics[width = 0.7\textwidth]
{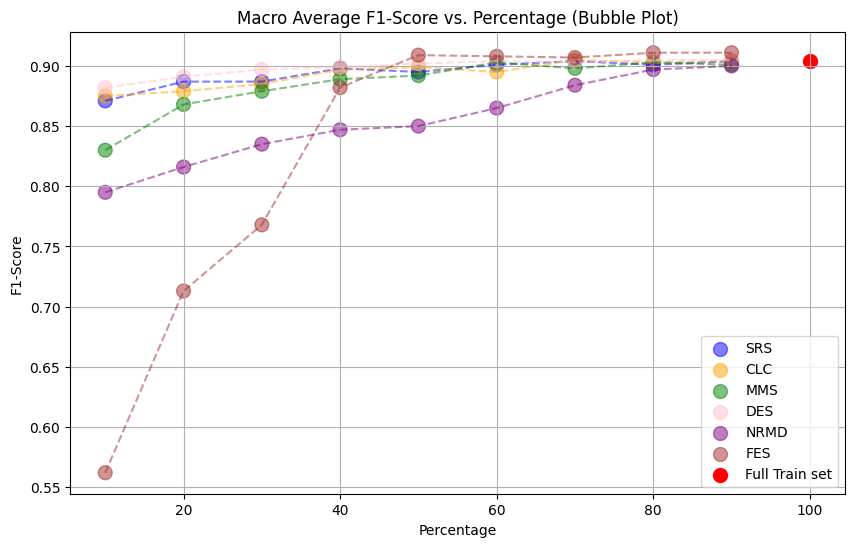}
        \caption[Collision: Reduction + macro average F1-score]{\label{collisionF1} Collision: Reduction + macro average F1-score}
        \end{figure}

The results on accuracy, macro average precision, macro average recall and macro average F1-score can be seen in Figures \ref{collisionAccuracy}, \ref{collisionPrecision}, \ref{collisionRecall} and \ref{collisionF1} respectively.
With respect to the accuracy, we can see that the model trained with the whole training dataset has an median success probability of $91\%$. In general, all the data reduction methods work very well for this dataset. In fact, there are many specific cases where the model obtained with a reduced dataset performs better on the test dataset than the one trained with the complete training dataset. We observe that when the reduction ratio is above $50\%$ the best performing method is FES, while in other cases it is DES. Most of the compared methods manage to maintain accuracy almost intact despite the significant reduction in training size. If we look at the results when we reduce the training dataset to $10\%$ of its size, the model trained after applying DES loses $1.8\%$ of accuracy, while the loss less than $3\%$ when CLC and SRS are applied. In all cases, this loss in accuracy is more or less linear for all methods except for FES. In this case, the accuracy remains stable while the reduction ratio is high, but it undergoes a drastic drop when a high percentage of examples is removed.

We observe a similar situation when analyzing the macro average precision. FES is the best method to preserve this metric (even improving the reference case) when the reduction ratio is greater than $50\%$, while for other ratios the best one is DES. 
macro average precision dropping as training data set size decreases
also appears to be approximately linear except for all the methods but FES, which suffers a significant drop when the reduction ratio is under $30\%$. 

The results that we get when we analyze the macro average recall are quite different. As we saw with accuracy and macro average precision, this metric is generally well preserved even if the reduction ratio is very low, although it suffers a very significant drop when FES is applied with a reduction ratio under $30\%$. But contrary to what we have seen for the previous metrics, no method clearly outperforms the rest in terms of macro average recall. Depending on the reduction ratio, the method that best preserves the macro average recall is one or another. All methods except NRMD have given the best median result for some of the chosen percentages.

All the general observations we have made when analyzing accuracy and macro average precision can also be seen for the macro average F1-score. In general, all reduction methods preserve this metric well, being FES the best performing method when the reduction ratio is higher than $50\%$ and DES otherwise. The drop in macro average F1-score is also noticeable when many examples are removed with the FES method, while this tendency is not as pronounced for the other data reduction methods.

Finally, we have found an interesting relationship between the $\varepsilon$-rep\-re\-sen\-ta\-ti\-ve\-ness of the reduced datasets and the macro average F1-score of the models trained with them. Given any reduction ratio $p=0.1,\cdots,0.9$, we have got the $\varepsilon$-rep\-re\-sen\-ta\-ti\-ve\-ness
and the macro average F1-score for each reduction method and each iteration in the experiment (in total there are $6 \text{ reduction methods}\times 10 \text{ iterations} = 60 \text{ pairs } (\varepsilon,\text{F1})$ for each $p$). We computed for each $p$ the Spearman's rank correlation coefficient \cite{spearman1904proof} of its respective cloud of 60 points to test if there exists a dependence between $\varepsilon$-representativeness and the macro average F1-score that can be described with a monotonic (always increasing or always decreasing) function. This coefficient is a real number $\rho \in [-1,1]$, where $\rho$ close to $1$ indicates a strong positive monotonic correlation, implying that as one variable increases, the other variable also increases, $\rho$ close to $-1$ indicates a strong negative monotonic correlation, and $\rho$ similar to $0$ indicates no monotonic correlation. We also compute the associated p-value to test if the correlation $\rho$ is significantly different from $0$. A $p$-value under a certain threshold (in our case  $0.05$, which is a standard choice) indicates that $\rho$ is likely not null, while a p-value above it suggests that the observed correlation might be coincidental and not due to a true dependence between both variables. We performed this statistical analysis independently for each $p$ to eliminate the possible effect that the reduction ratio could have if we used all the possible pairs $(\varepsilon,\text{F1})$ altogether.

\begin{table}[H]
    \centering
    \resizebox{0.8\columnwidth}{!}{
    \begin{tabular}{|l|c|c|c|c|c|c|c|c|c|}
    \hline
    \cellcolor{gray!20} & \cellcolor{gray!20} \textbf{10\%} & \cellcolor{gray!20} \textbf{20\%} & \cellcolor{gray!20} \textbf{30\%} & \cellcolor{gray!20} \textbf{40\%} & \cellcolor{gray!20} \textbf{50\%} & \cellcolor{gray!20} \textbf{60\%} & \cellcolor{gray!20} \textbf{70\%} & \cellcolor{gray!20} \textbf{80\%} & \cellcolor{gray!20} \textbf{90\%} \\
    \hline
    \cellcolor{gray!20} \textbf{Spearman's $\rho$} & \cellcolor{green!10}-0.38& \cellcolor{green!10}-0.43& \cellcolor{green!10}-0.42& \cellcolor{green!10}-0.39& -0.22& -0.15& -0.19& -0.07& -0.14
\\
    \hline
    \cellcolor{gray!20} \textbf{$p$-value} & \cellcolor{green!10}0.0& \cellcolor{green!10}0.0& \cellcolor{green!10}0.0& \cellcolor{green!10}0.0& 0.1& 0.24& 0.14& 0.58& 0.3
\\
    \hline
    \end{tabular}}
    \caption[Collision: Correlation between $\varepsilon$-Representativeness and macro average F1-score]{Collision: Correlation between $\varepsilon$-representativeness and macro average F1-score. This table displays the non-parametric Spearman correlation coefficient and its p-value. We have marked in green the columns with a significative correlation setting a significance level of 5\%, that is with a $p$-value less or equal than 0.05}
    \label{tab:col_eps_f1}
\end{table}

The results that we got can be seen in Table \ref{tab:col_eps_f1}. All the computed $\rho$ values are negative, although they are only significantly different from zero when the reduction ratio is below $40\%$. That indicates that, when data reduction methods remove a large number of examples, the best-performing models are those trained with the reduced datasets that best preserve the $\varepsilon$-representativeness of the entire training set. In few words, when we reduce the Collision dataset with a small reduction percentage, the smaller the $\varepsilon$ value, the better the model will perform.

\paragraph*{Dry Bean  Dataset}

\begin{figure}[H]
        \centering
    \includegraphics[width = 0.7\textwidth]
{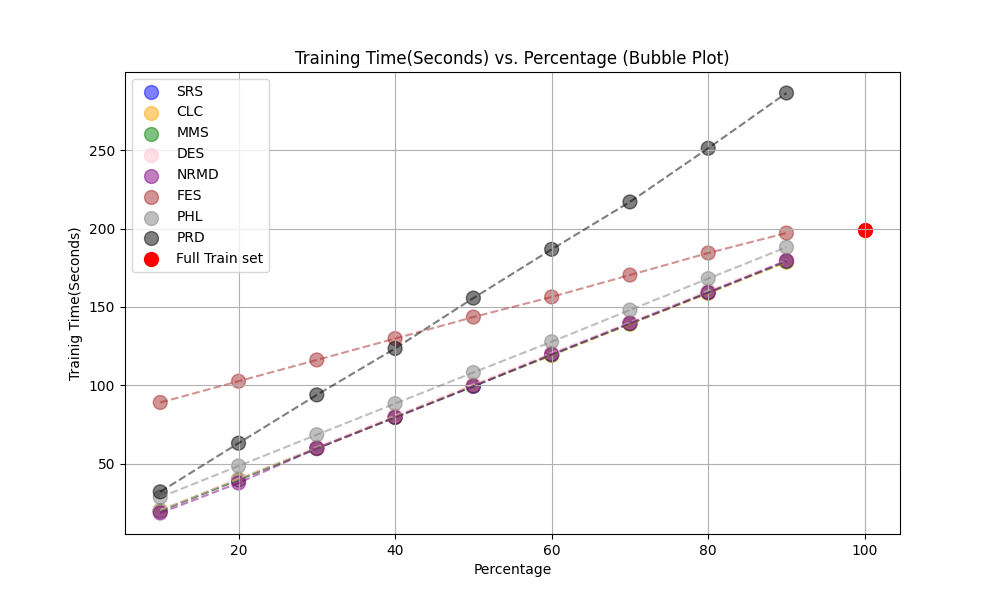}
        \caption[Dry Bean: Reduction + Training time]{\label{DryBeanTT} Dry Bean: Reduction + training time}
        \end{figure}
\begin{figure}[H]
        \centering
    \includegraphics[width = 0.7\textwidth]
{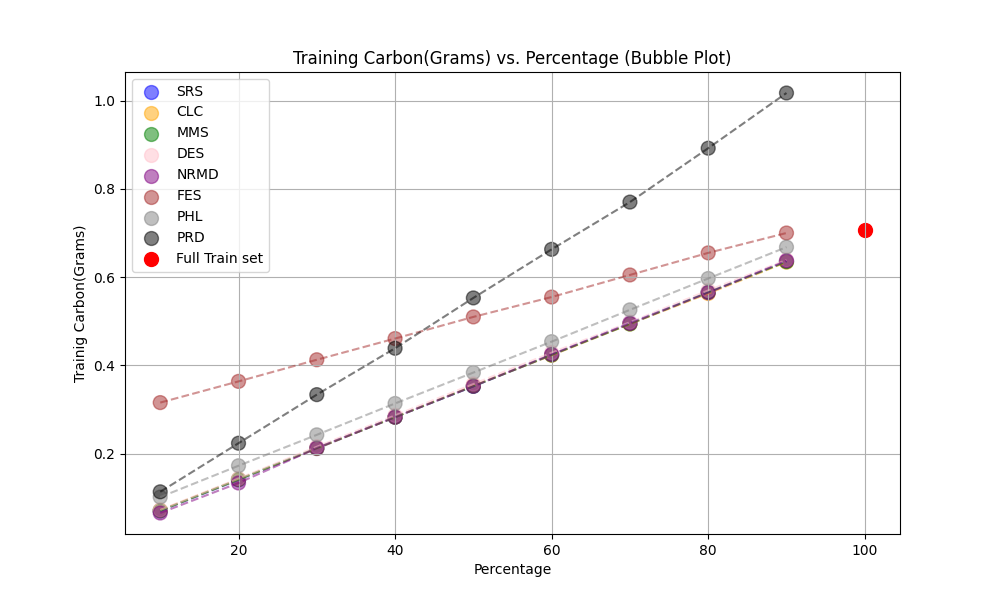}
        \caption[Dry Bean: Reduction + Training carbon]{\label{DrybeanTC} Dry Bean: Reduction + training carbon}
        \end{figure}

The median results for the computing time and carbon emission can be seen in Figures \ref{DryBeanTT} and \ref{DrybeanTC}. There is also a proportional relation between the computation time and the carbon emission in this experiment since each minute of computations emitted approximately $0.21$ g of $\text{CO}_2$ into the atmosphere. As happened with the Collision dataset, the use of data reduction methods prior to network training helped to reduce the computation time and the carbon emission of the model building with respect to the reference case. The only exception to this rule is when we apply PRD reduction with a reduction ratio greater than $70\%$ (see in Figures \ref{DryBeanTT} and \ref{DrybeanTC}). In that situation, both the computation time and the carbon emission of 
reduction and 
training exceed those of the reference case. There is no reduction method that runs faster than all the others for this dataset. SRS, MMS, DES, NRMD and even CLC reduction(all of these are overlappingby NRMD in these figures) , which was the slowest method for the Collision dataset, run equally fast for the Dry Bean dataset. They hardly need any time to reduce the training data set, so almost all the measured time and, therefore, the carbon emission correspond to the network training.

\begin{figure}
        \centering
    \includegraphics[width = 0.7\textwidth]
{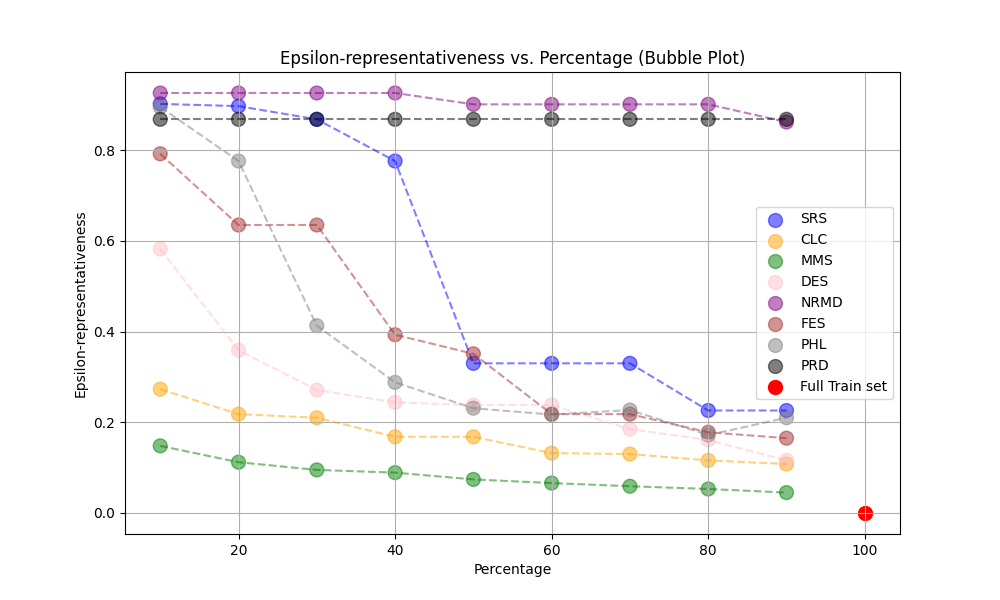}
        \caption[Dry Bean: Reduction + Epsilon]{\label{DrybeanEpsilon} Dry Bean: Reduction + Epsilon}
        \end{figure}

The median values on the $\varepsilon$-representativeness, which can be seen in Figure \ref{DrybeanEpsilon}, show us similar results to those observed with the Collision dataset. MMS is still the best data reduction method to preserve the $\varepsilon$-representativeness with respect to the full training dataset, being CLC the second best option. On the contrary, NRMD and PRD are the two methods that generally give the less $\varepsilon$-representative reduced datasets.

\begin{figure}
        \centering
    \includegraphics[width = 0.7\textwidth]
{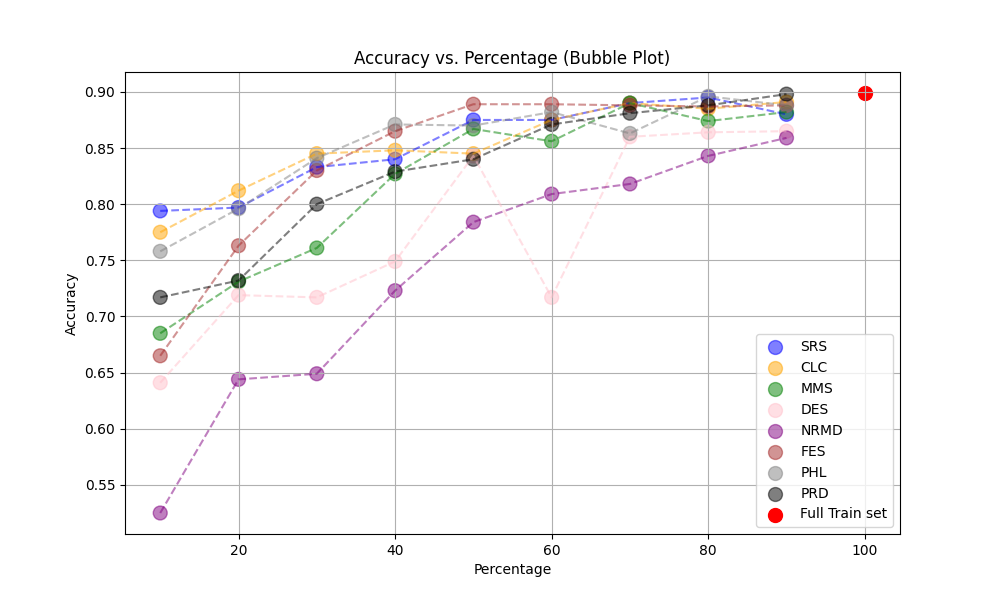}
        \caption[Dry Bean: Reduction + Accuracy]{\label{DrybeanAccuracy} Dry Bean: Reduction + Accuracy}
        \end{figure}

\begin{figure}
        \centering
    \includegraphics[width = 0.7\textwidth]
{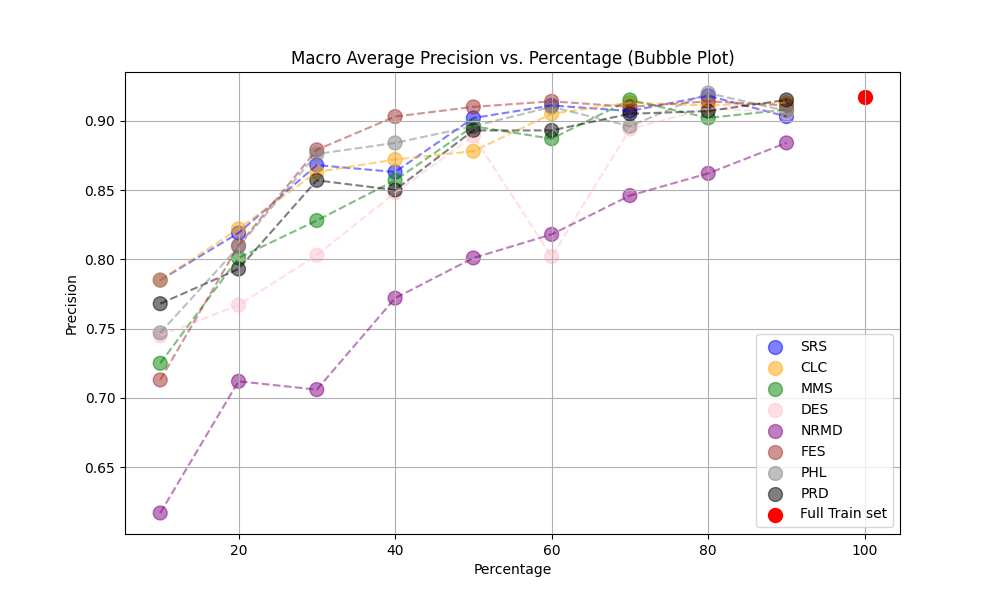}
        \caption[Dry Bean: Reduction + Precision]{\label{DrybeanPrecision} Dry Bean: Reduction + Precision}
        \end{figure}

\begin{figure}
        \centering
    \includegraphics[width = 0.7\textwidth]
{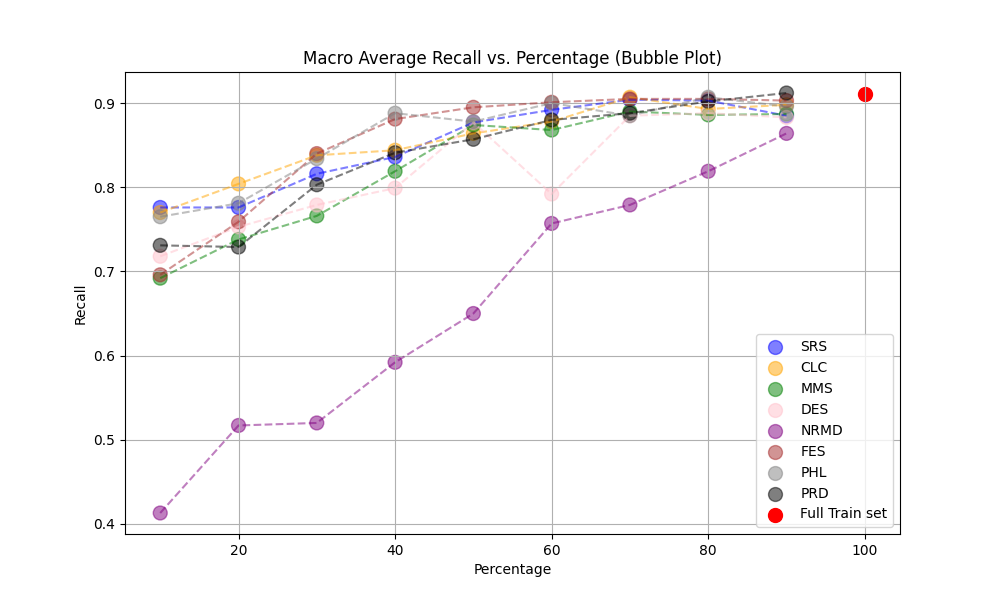}
        \caption[Dry Bean: Reduction + Recall]{\label{DrybeanRecall} Dry Bean: Reduction + Recall}
        \end{figure}

\begin{figure}
        \centering
    \includegraphics[width = 0.7\textwidth]
{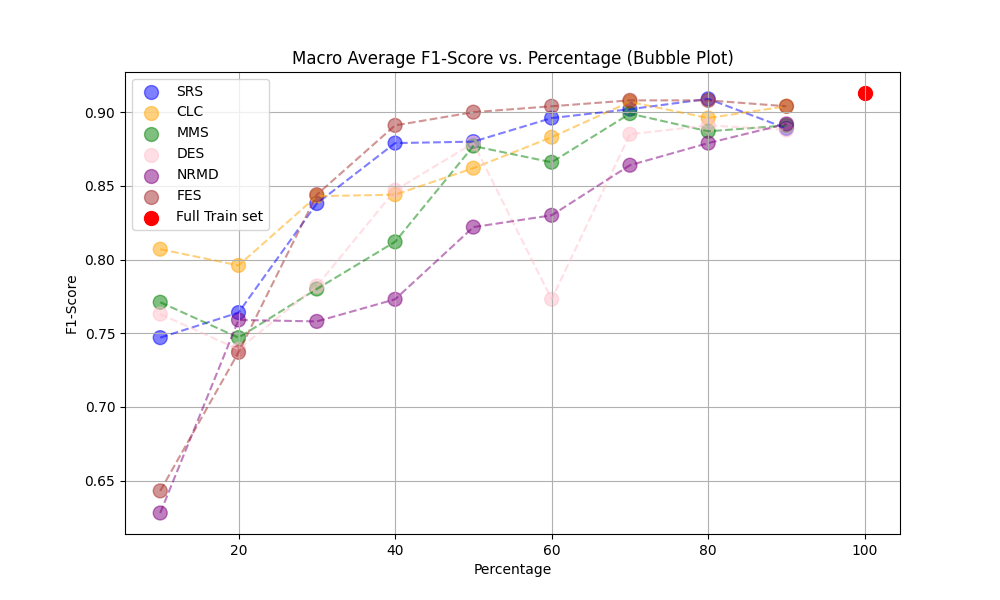}
        \caption[Dry Bean: Reduction + F1-Score]{\label{DrybeanF1Score} Dry Bean: Reduction + F1-Score}
        \end{figure}

The median results on accuracy, macro average precision, macro average recall and macro average F1-score for the Dry Bean dataset can be seen in Figures \ref{DrybeanAccuracy}, \ref{DrybeanPrecision}, \ref{DrybeanRecall} and \ref{DrybeanF1Score} respectively.
In this experiment, the reference case has a median accuracy in the test dataset of $89.9\%$ and, contrary to the Collision dataset, no model trained on a reduced dataset improves this value. No reduction method outperforms the others for every reduction ratio in terms of accuracy. 
What we observe is that, while, in general, the accuracy is well preserved when the reduction ratio is high, it suffers a drastic drop when many examples are removed from the training dataset. If we wanted to lose at most $5\%$ of accuracy (i.e., have at least $84.9\%$) we would have to select at least $40\%$ of the data, and not all reduction methods would guarantee that maximum accuracy loss. 

With respect to macro average precision, we also find that no method is clearly better than the rest for all reduction ratios, although FES seems to dominate the statistics for the central ratios ($30\% \leq p \leq 60\%$). Here, the drop in the metric as the reduction ratio decreases is not as pronounced as it is for accuracy. This could be because the macro average precision is a more robust measure as it is less influenced by the larger classes.

We can see similar results when analyzing the details on macro average recall. No method outperforms all the rest, and SRS, PRD, CLC, PHL and FES give the best result for at least one reduction ratio.

Finally, when analyzing the macro average F1-score, we observe that four methods (PRD, CLC, PHL and FES) give the best score for some reduction ratio, but FES seems to give the best performing reduced datasets for $30\% \leq p \leq 70\%$. As we said when we analyzed the accuracy, it is not possible to extract a reduced dataset with $30\%$ of its size or less without losing more than $5\%$ of macro average F1-score. 

\paragraph*{Description of data structure for accessing the final results}

The results of both experiments were saved in a Python dictionary with a nested structure, which we will call the \textit{results dictionary}. On the first level, the result dictionary is composed of 10 dictionaries containing 
a dictionary for each iteration.
The dictionary of each iteration has 
a dictionary for each data reduction method used.
The dictionary of each data reduction method contains an item for each reduction ratio $p = 0.1, \cdots, 1.0$, and each one of them is a dictionary containing all the results obtained for that data reduction method and the $p$ value in that iteration. The results obtained when the training dataset is not reduced are saved in the item with key $p=1.0$. Finally, the dictionary associated with a specific iteration, data reduction method, and reduction ratio, contains a key for the following metrics: 

\begin{itemize}
    \item \textbf{time:} To store the computing time in seconds of reduction and training (only training when $p=1.0$)
    \item \textbf{carbon:} To store the carbon emission in kg of $\text{CO}_2$ of reduction and training (only training when $p=1.0$)
    \item \textbf{epsilon:} To store the $\varepsilon$-representativeness of $\dd_{\text{train},R}$ with respect to $\dd_{\text{train}}$
    \item \textbf{acc:} To store the accuracy of the model over $\dd_{\text{test}}$
    \item For each class $k$:
    \begin{itemize}
        \item \textbf{pre\_k:} To store the model precision for class $k$ over $\dd_{\text{test}}$
        \item \textbf{rec\_k:} To store the model recall for class $k$ over $\dd_{\text{test}}$
        \item \textbf{f1\_k:} To store the model F1-score for class $k$ over $\dd_{\text{test}}$
    \end{itemize}
    \item \textbf{pre\_avg:} To store the model macro average precision for class $k$ over $\dd_{\text{test}}$
    \item \textbf{rec\_avg:} To store the model macro average recall for class $k$ over $\dd_{\text{test}}$
    \item \textbf{f1\_avg:} To store the model macro average F1-score for class $k$ over $\dd_{\text{test}}$
\end{itemize}

Once we have this results dictionary, the next step is to summarize the information of all the iterations in a more simple dictionary. This object, which we will call the \textit{median results dictionary}, has the same structure as the entry that we got for each iteration in the results dictionary. For each data reduction method and each reduction ratio, the entry of a specific metric is the median value of the 10 metrics obtained during the 10 different iterations of the experiment. This way we can obtain a more stable representation of the performance of each method across iterations, mitigating the potential influence of outliers or variability in each individual run. All the Figures that we have seen in this subsection present the metrics from the median results dictionary.

\subsection{Experiments for Object Detection}
    In this subsection, we describe 
    the methodology we have used in our experiments to extend data reduction techniques to images. Observe that we need to adapt the methodology depending on the type of data reduction method. Later, we present the datasets used for the experiments done on object detection, including the parameter settings and the setup, and finally, we discuss the results that we have obtained.
    \subsubsection{Methodology}\label{sec:methodology}

\begin{figure}
        \centering
        \includegraphics[width=\linewidth]{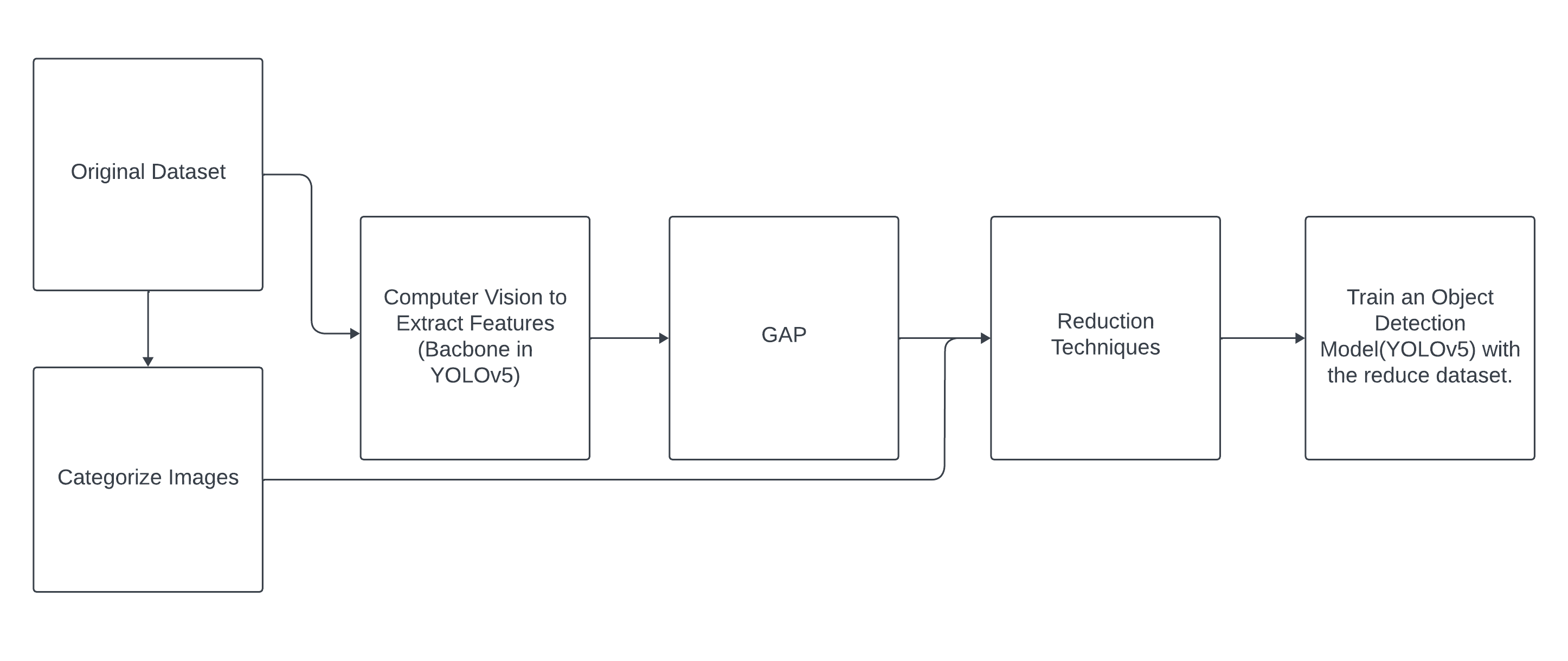}
        \caption[Proposed Methodology to Apply Data Reduction Techniques on Images]{\label{methodologyDRImages} Diagram of the workflow for the proposed methodology to apply data reduction techniques on images dataset.}
        \end{figure}

        The proposed methodology for the data-dependent methods, illustrated in Figure~\ref{methodologyDRImages},
        consists of the following five steps:
        
        \begin{enumerate}

        \item \textbf{Feature extraction}: The objective of this stage is to convert the raw pixel values of images into a set of meaningful and concise features that capture pertinent information. These features should empower the model to distinguish between different patterns, objects, or structures within the images. This process entails utilizing a computer vision model, 
        to extract features from all the images in the training set. In our case, we used a pre-trained YOLOv5 model on the COCO dataset (Common Objects in Context) \cite{lin2015microsoft}, a widely used collection in computer vision. The COCO dataset, consisting of approximately 330,000 annotated images with object location and category information, is one of the largest and most diverse datasets available for object detection and segmentation tasks. Utilizing pre-trained models on COCO proves advantageous because of its scale and diversity. This pre-training allows models to learn generic features and representations from a vast array of real-world images, enhancing their ability to generalize across various downstream tasks. This approach can lead to improved performance and efficiency when fine-tuning or adapting these models to specific applications. 

        \item \textbf{Categorizing images}:  In this step, we categorize each image based on the objects present in them. This categorization is essential for applying data reduction techniques, as explained in section \ref{sec:datasetsOD}, where we detail how each dataset is categorized.
        
        \item \textbf{Global Average Pooling}: Regarding the output of Step 1, we apply Global Average Pooling \cite{gholamalinezhad2020pooling, lin2014network}to the output of the last layer of the backbone, in order to transform the features maps into an n-dimensional vector representing their extracted features. Subsequently, these feature vectors can be used to calculate distances or similarities between images, and reduction methods can be applied to them.

        \item \textbf{Applying data reduction technique}: Reduction techniques are applied to decrease the amount of samples in the dataset with a specified reduction rate on the matrix produced in Step 3, comprising x images and n dimensions, along with the labels from Step 2.

        \item \textbf{Fine tuning with the reduced dataset}: This step allows us to assess whether satisfactory performance is achieved, potentially maintaining the same level as with the complete training set. Performance evaluation is conducted on the test set using YOLOv5 pre-trained on the COCO dataset, with the backbone frozen. In this context, "fine-tuning with a specific part frozen" implies that some of the model’s parameters are kept fixed during the training process on the new task. This approach leverages prior knowledge gained during initial training, enabling more efficient adaptation to the new task without completely discarding previously learned information.
        \end{enumerate}
        
\begin{figure}
        \centering
        \includegraphics[width=\textwidth]{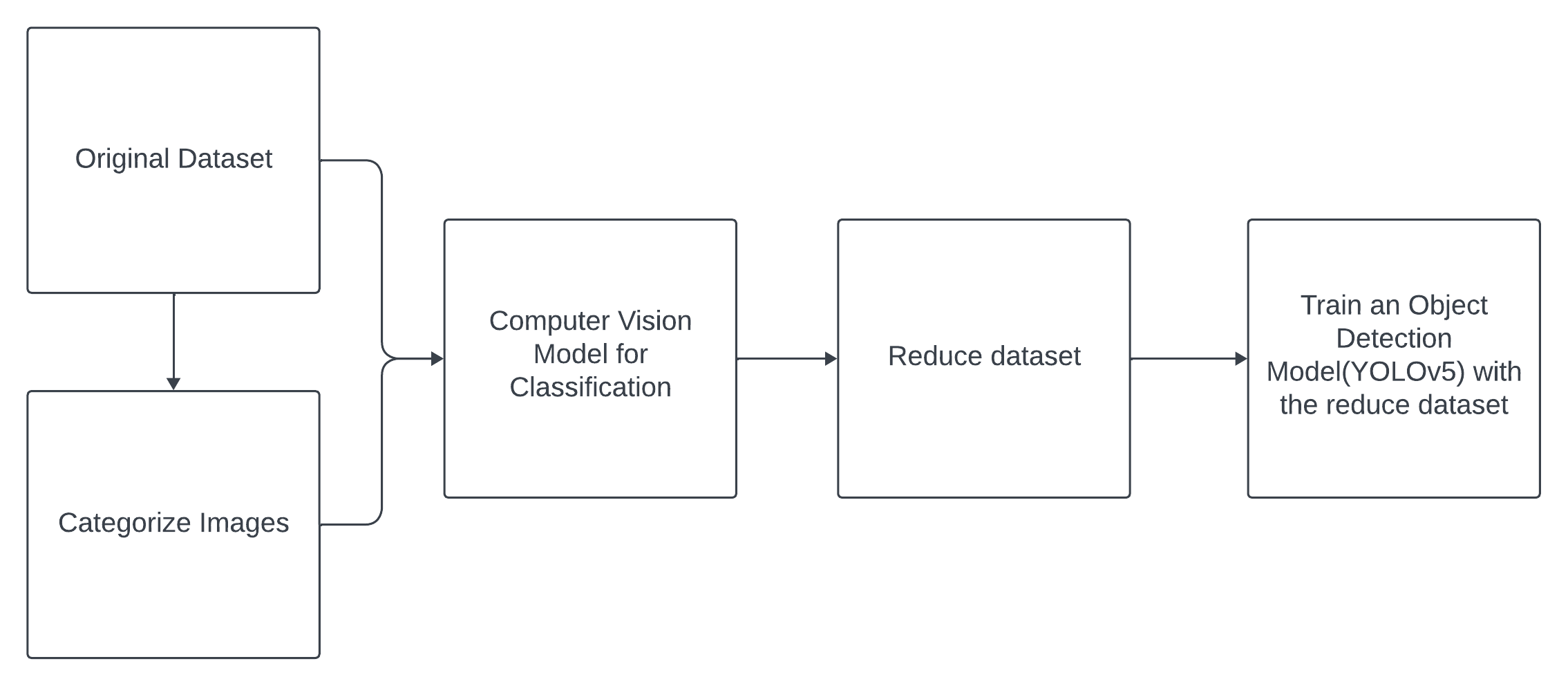}
        \caption[Proposed Methodology to Apply Wrapper Data Reduction Techniques on Images]{\label{methodologyDRWRImages} Diagram of the workflow for the proposed methodology to apply wrapper methods.}
        \end{figure}

        When employing wrapper methods, we must adopt a slightly different method\-ol\-o\-gy than the one previously described (see Figure \ref{methodologyDRWRImages}). Initially, we must categorize the images, similar to the preceding methodology, as these images are intended for object detection and lack specific labels, instead featuring multiple elements within them. Subsequently, we create a straightforward classification model, incorporating the reduction technique during training to yield the reduced dataset. Finally, we train the YOLOv5 detection model using the reduced dataset, aligning with the objective of Step 5 in the aforementioned methodology.

        Additionally, to apply the CLC method to images, we introduce a slight modification by employing KMeans \cite{IKOTUN2023178} on $X$ with $c$ (the number of classes) clusters. Then, we determine the number of samples closest (based on the Euclidean distance) to each centroid, resulting in our reduced dataset $\dd_R$ guided by the specified reduction rate. This adjustment is necessary because the centroids generated by the KMeans method, derived from the representations obtained by our image methodology, do not correspond to specific images from our dataset. Consequently, they do not convey information about what we genuinely aim to detect and localize. This modification is called Representative KMeans (RKM).
        
        \subsubsection{Datasets for Object Detection}\label{sec:datasetsOD}
        
            \paragraph*{Roboflow}
            The dataset Roboflow\footnote{\url{https://universe.roboflow.com/2458761304-qq-com/wheelchair-detection}} \cite{wheelchair-detection_dataset} comprises 514 RGB images, each 416 pixels in both height and width. These images feature pedestrians and people in wheelchairs. The training dataset comprises 463 RGB images, in which a total of 499 pedestrians (annotated as P) and 616 wheelchair users (annotated as W) appear. The test dataset is composed of 51 RGB images, in which a total of 55 pedestrians and 65 wheelchair users appear. 
            To label this dataset, we stick to the following criteria:
\begin{center}
    \begin{tabular}{|c|c|c|}
\hline
\cellcolor{gray!20}\textbf{Number of P}& \cellcolor{gray!20}\textbf{Number of W}& \cellcolor{gray!20}\textbf{label}\\ \hline
0& 1& 0\\ \hline
$\ge 1$& 0& 1\\ \hline
$\ge 1$& $\ge 1$& 2\\ \hline
0& $\ge 2$& 3\\ \hline
\end{tabular}
\end{center}

            \paragraph*{Mobility Aid}
            The dataset Mobility Aid \footnote{\url{http://mobility-aids.informatik.uni-freiburg.de/}} \cite{vasquez17ecmr} is composed of 17079 RGB images, 10961 are part of the training dataset and the 6118 remaining are part of the test dataset. In this dataset we can find five types of objects: pedestrians (8371, it will be annotated as P), wheelchair users (6458, it will be annotated as W), person pushing a wheelchair (3323, it will be annotated as PW), person with crutches (5374, it will be annotated as C) and person with a walking-frame (7649, it will be annotated as WF). The test dataset is composed of 6208 P, 1993 W, 782 PW, 1883 C, and 2174 WF.
            To label this dataset, we stick to the following criteria:
            \begin{center}
                \begin{tabular}{|c|c|c|}
\hline
\cellcolor{gray!20}\textbf{Number of P/PW}&\cellcolor{gray!20}\textbf{ Number of W/C/WF}& \cellcolor{gray!20}\textbf{label}\\ \hline
1& 0& 1\\ \hline
0& 1& 2\\ \hline
$\ge 1$& 0& 3\\ \hline
0& $\ge 1$& 4\\ \hline
$\ge 1$& $\ge 1$& 5\\ \hline
\end{tabular}
            \end{center}

        \subsubsection{Parameter Setting}
        
        For the fine-tuning of YOLOv5, we maintain the backbone frozen (pretrained on the COCO dataset) while training the remaining parts of the model. We configure the training with 100 epochs for the Roboflow dataset and 50 epochs for the Mobility Aid dataset. The batch size is set at 16, and the image size is fixed at 640. We employ the SGD optimizer, and the learning rate is set to 0.01 for both datasets. The number of classes (nc) is adjusted based on the specific dataset, such as 2 for the Roboflow dataset and 5 for the Mobility Aid dataset.
        
        \subsubsection{Experiments Setup}

        We utilized Python 3.9 and PyTorch \cite{Pytorch} on Ubuntu 20.04 to conduct our experiments. The training phase was executed on an NVIDIA QUADRO RTX 4000 with 8 GB of RAM and an Intel Xeon Silver 4210 preprocessor.
        Data partitioning followed the default specifications for each dataset.

        To assess the performance difference between the full training set and the reduced set, we conducted training five times for the reduced training dataset. We calculated the arithmetic mean and standard deviation of the results for the test set to gauge the performance of each reduction method. For the Roboflow dataset, given its limited sample size, we applied rate reductions of 50\% and 75\%. In contrast, for the Mobility Aid dataset with a larger sample size, we implemented rate reductions of 75\% and 90\%.

\subsubsection{Comparison Metrics}

We use five metrics to assess the performance of training YOLOv5 with the complete dataset compared to training it with different reduction methods and reduction rates. Initially, we employ three performance metrics: precision, recall, and mean average precision setting a confidence threshold of 0.5 ($mAP$@0.5), which means that only detections with a confidence of 50\% or more are considered. These metrics are computed individually for each class and globally by averaging the values across all classes. Additionally, we consider the time required for data reduction, measured in seconds for each reduction method, and the model fine-tuning time. Our primary goal with these metrics is to determine whether we can maintain similar performance while considering the time saved.

Furthermore, we calculate $\varepsilon$-representativeness to gauge how well the reduced dataset $\dd_R$ represents the original dataset $\dd$. 
We also compute carbon emissions during both the data reduction process 
and during fine-tuning.

\subsubsection{Results and Discussion}

The source code to implement data reduction techniques in these datasets is available at \url{https://github.com/Cimagroup/ExperimentsOD-SurveyGreenAI} \cite{Victor_Toscano_Duran_Repository_experiments_Survey_2023} in the folder ObjectDetection.

\paragraph*{Roboflow Dataset}

In our initial experiment, we compared the performance of fine-tuning with the complete training dataset against training with a reduced dataset using various reduction methods. Table \ref{tab:B1-50} displays the reduction time, $\varepsilon$-representativeness of the $\dd_r$, training time, model performance, and $\text{CO}_2$ emissions during the training and reduction phases at a 50\% reduction rate. We can observe that, despite training with only 50\% of the data, we maintained performance comparable to using the entire dataset. Additionally, we reduced the training time by approximately 40\%, decreasing from 9 minutes and 20 seconds to around 5 minutes and 45 seconds with 50\% of the samples. This time reduction was accompanied by a similar decrease (about 40\%) in $\text{CO}_2$ emissions throughout the process. The emission during the application of the reduction methods was small compared to the training time. Notably, effective methods in this scenario include SRS, DES, MMS, RKM, and FES. On the contrary, NRMD and PHL perform worse, with a slight loss of performance. Data reduction times were generally light, except for FES, which exhibited excessive duration compared to other methods. Conversely, a lower $\varepsilon$-representativeness did not seem decisive for performance improvement or degradation.

Table \ref{tab:B1-75} presents the same analysis with a 75\% reduction rate, where the overall training time was reduced from 9.5 minutes to approximately 4 minutes, a 60\% increase in speed. Data reduction time remained insignificant, with some methods displaying longer computation times, such as FES. Despite a slight performance reduction across all metrics with all methods, SRS, MMS, RKMEANS, and PHL emerged as more resilient options. In particular, SRS is the one that best maintains precision and $mAP$. On the contrary, NRMD showed the most significant performance loss. A lower $\varepsilon$-representativeness did not appear to be a determining factor for better or worse performance. $\text{CO}_2$ emissions were also reduced by 60\%. On average, the overall $mAP$ for reduction methods saw only a 3\% reduction compared to the substantial computational time and $\text{CO}_2$ emission savings of 60\%.

\begin{figure}[H]
        \centering
        \includegraphics[width=\textwidth]{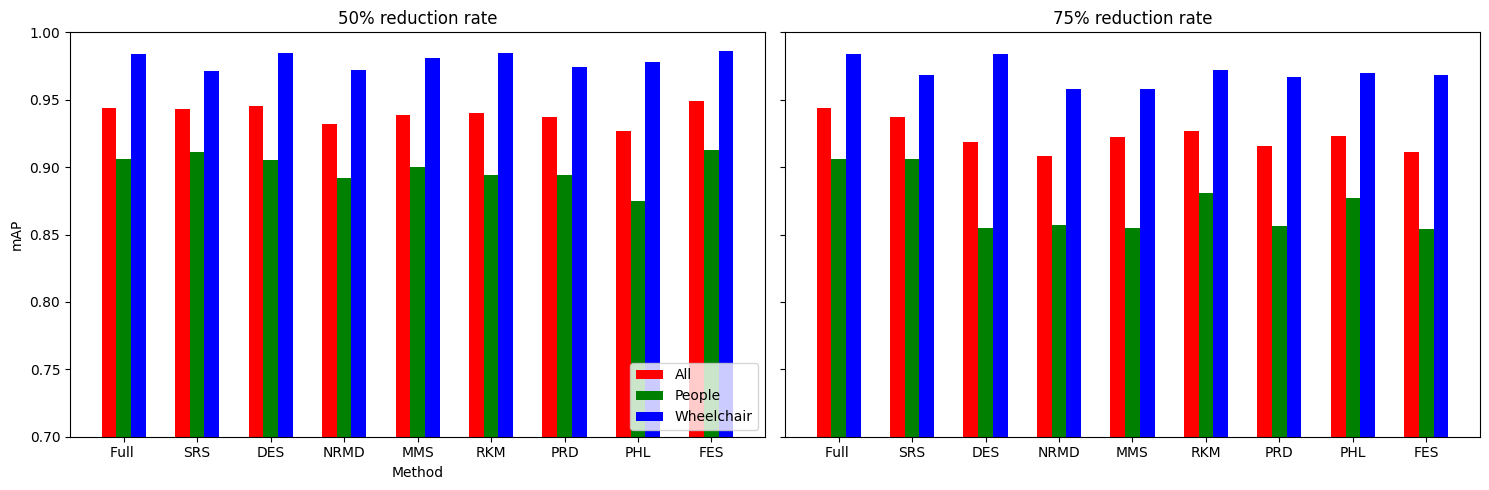}
        \caption[$mAP$ Values on Roboflow Dataset]{\label{$mAP$Roboflow} $mAP$ values on Roboflow dataset when using a 50\% 
        reduction rate (first column) and when using a 75 percent reduction rate (second column).}
        \end{figure}

In Figure \ref{$mAP$Roboflow}, the mean $mAP$ values for each category and method, along with the full dataset, illustrate the best-performing methods. At a 50\% reduction rate, performance is nearly maintained, while at a 75\% reduction, some performance loss is evident. SRS stands out as the most effective method. Additionally, a general improvement in accuracy is observed for Wheelchairs compared to People, potentially attributed to a slight imbalance in the dataset between the two categories. We only present the $mAP$ figure (Figure~\ref{$mAP$Roboflow}) because it is the most comprehensive performance metric for object detection evaluation.

Based on the outcomes obtained from this dataset, we can affirm that employing reduction methods within the proposed methodology, followed by fine-tuning YOLOv5 for object detection, led to a substantial reduction in $\text{CO}_2$ emissions and computation time. Importantly, this reduction did not adversely affect the model's performance in localization and object detection tasks.

\begin{table}[H]
    \centering
    \resizebox{0.9\columnwidth}{!}{
    \begin{tabular}{|c|c|c|c|c|c|c|c|}
        \hline
        \cellcolor{gray!20}\textbf{Method}& \cellcolor{gray!20}\textbf{R Time(s)}& \cellcolor{gray!20}\textbf{$\varepsilon$}& \cellcolor{gray!20}\textbf{FT Time}& \cellcolor{gray!20}\textbf{Precision}& \cellcolor{gray!20}\textbf{Recall}& \cellcolor{gray!20}\textbf{$mAP$@.5}& \cellcolor{gray!20}\textbf{CO}$_2$\textbf{(g)}\\
        \hline
        \rowcolor{red!10}
        - & - & - & 9m 19s & \begin{tabular}{@{}c@{}}A: 0.951$\pm$0.001 \\ P: 0.926$\pm$0.022 \\ W: 0.976$\pm$0.015\end{tabular} & \begin{tabular}{@{}c@{}}A: 0.897$\pm$0.019 \\ P: 0.832$\pm$0.031 \\ W: 0.96$\pm$0.018\end{tabular} & \begin{tabular}{@{}c@{}}A: 0.944$\pm$0.009 \\ P: 0.906$\pm$0.014\\ W :0.984$\pm$0.005\end{tabular} & 5.5 \\
        \hline
        SRS & \cellcolor{green!10} 0.002 & 2.58 & 5m 44s & \begin{tabular}{@{}c@{}}A: 0.945$\pm$0.021 \\ P: 0.921 $\pm$0.035 \\ W: 0.97$\pm$0.011\end{tabular} & \begin{tabular}{@{}c@{}}A:0.897 $\pm$0.019  \\ P: 0.836$\pm$0.03  \\ W: 0.958$\pm$0.015\end{tabular} & \begin{tabular}{@{}c@{}}A: 0.943$\pm$0.01 \\ P: 0.911$\pm$0.014\\ W: 0.971$\pm$0.005\end{tabular} & 0+3.3 \\
        \hline
        DES & 0.29 & 3.12 & 5m 44s  & \begin{tabular}{@{}c@{}}A: 0.94$\pm$0.001 \\ P: 0.925$\pm$0.015 \\ W: 0.95$\pm$0.015\end{tabular} & \begin{tabular}{@{}c@{}}A: 0.885$\pm$0.015  \\ P: 0.795$\pm$0.03  \\ W: 0.975$\pm$0.015 \end{tabular} &  \begin{tabular}{@{}c@{}}A: 0.945$\pm$0.005 \\ P:0.905$\pm$0.005\\ W: 0.985$\pm$0.005\end{tabular} & 0.002+3.3 \\
        \hline
        NRMD & 0.09 & 2.27 & 5m 47s & \begin{tabular}{@{}c@{}}A: 0.925$\pm$0.015 \\ P: 0.905$\pm$0.02 \\ W: 0.945$\pm$0.02\end{tabular} & \begin{tabular}{@{}c@{}}A: 0.89$\pm$0.016  \\ P: 0.82$\pm$0.014  \\ W: 0.956$\pm$0.019\end{tabular} & \begin{tabular}{@{}c@{}}A: 0.932$\pm$0.009 \\ P: 0.892$\pm$0.016\\ W : 0.972$\pm$0.007\end{tabular} & 0.001+3.3 \\
        \hline
        MMS & 0.09 & 1.99 & 5m 46s & \cellcolor{green!10} \begin{tabular}{@{}c@{}}A: 0.951$\pm$0.01 \\ P: 0.921$\pm$0.017  \\ W: 0.981 $\pm$0.005\end{tabular} & \begin{tabular}{@{}c@{}}A: 0.894$\pm$0.015  \\ P: 0.821$\pm$0.029  \\ W: 0.967$\pm$0.011)\end{tabular} & \begin{tabular}{@{}c@{}}A: 0.939$\pm$0.006 \\ P:0.9$\pm$0.011\\ W: 0.981$\pm$0.006 \end{tabular} & 0.0004+3.26 \\ 
        \hline
        RKM & 1.26 & \cellcolor{green!10} 1.25 & 5m 45s  & \begin{tabular}{@{}c@{}}A: 0.948$\pm$0.005 \\ P:0.907$\pm$0.015 \\ W: 0.99 $\pm$0.009\end{tabular} & \begin{tabular}{@{}c@{}}A: 0.895$\pm$0.017  \\ P: 0.819$\pm$0.032  \\ W: 0.971$\pm$0.008 \end{tabular} & \begin{tabular}{@{}c@{}}A: 0.94$\pm$0 \\ P: 0.894$\pm$0.003\\ W: 0.985$\pm$0.002\end{tabular} & 0.005+3.3 \\
        \hline
        PRD & 0.92 & 1.67 & 5m 44s & \begin{tabular}{@{}c@{}}A: 0.944$\pm$0.02 \\ P: 0.916$\pm$0.04 \\ W: 0.97$\pm$0.008\end{tabular} & \begin{tabular}{@{}c@{}}A: 0.89$\pm$0.015 \\ P: 0.814$\pm$0.025  \\ W: 0.965$\pm$0.008\end{tabular} & \begin{tabular}{@{}c@{}}A: 0.937$\pm$0.007 \\ P: 0.894$\pm$0.015 \\ W: 0.974$\pm$0.008\end{tabular} & 0.002+3.29 \\
        \hline
        PHL & 0.64 & 2.85 &  5m 38s & \begin{tabular}{@{}c@{}}A: 0.942$\pm$0.028 \\ P: 0.9$\pm$0.045 \\ W: 0.982$\pm$0.014\end{tabular} & \begin{tabular}{@{}c@{}}A: 0.863$\pm$0.21  \\ P: 0.773$\pm$0.041  \\ W: 0.954$\pm$0.01\end{tabular} & \begin{tabular}{@{}c@{}}A: 0.927$\pm$0.006 \\ P: 0.875$\pm$0.017\\ W: 0.978$\pm$0.011\end{tabular} & \cellcolor{green!10} 0.004+3.24 \\
        \hline
        FES & 8.14 & 2.28 & 5m 39s & \begin{tabular}{@{}c@{}}A:0.921$\pm$0.013  \\ P: 0.871$\pm$0.026 \\ W: 0.972$\pm$0.017 \end{tabular} & \cellcolor{green!10} \begin{tabular}{@{}c@{}}A: 0.903$\pm$0.011   \\ P: 0.844$\pm$0.018  \\ W: 0.962$\pm$0.006 \end{tabular} & \cellcolor{green!10} \begin{tabular}{@{}c@{}}A: 0.948$\pm$0.004 \\ P: 0.913$\pm$0.009 \\ W: 0.986$\pm$0.003 \end{tabular} &  0.08+3.24 \\
        \hline
    \end{tabular}}
    \caption[Table Results for Roboflow Dataset and 50\% Reduction Rate]{Table results for Roboflow dataset and 50\% reduction rate. The 'Precision', 'Recall' and '$mAP$@.5' columns display mean and standard deviation values for the specified variables. The '$\text{CO}_2(g)$' column indicates the grams of $\text{CO}_2$ emitted during the application of the reduction method and during the fine-tuning. The 'R Time(s)' column shows the time in seconds for data reduction, while the 'FT Time' column displays the time spent on fine-tuning the model. We have highlighted in red the values obtained during fine-tuning with the complete dataset,which serve as the reference. Additionally, in green, we highlight the best reduction method for each metric.}
    \label{tab:B1-50}
\end{table}

\begin{table}[H]
    \centering
    \resizebox{0.9\columnwidth}{!}{
    \begin{tabular}{|c|c|c|c|c|c|c|c|}
        \hline
        \cellcolor{gray!20}\textbf{Method}& \cellcolor{gray!20}\textbf{R Time(s)}& \cellcolor{gray!20}\textbf{$\varepsilon$}& \cellcolor{gray!20}\textbf{FT Time}& \cellcolor{gray!20}\textbf{Precision}& \cellcolor{gray!20}\textbf{Recall}& \cellcolor{gray!20}\textbf{$mAP$@.5}& \cellcolor{gray!20}\textbf{CO}$_2$\textbf{(g)}\\
        \hline
        \rowcolor{red!10}
        - & - & - & 9m 19s & \begin{tabular}{@{}c@{}}A: 0.951$\pm$0.001 \\ P: 0.926$\pm$0.022 \\ W: 0.976$\pm$0.015\end{tabular} & \begin{tabular}{@{}c@{}}A: 0.897$\pm$0.019 \\ P: 0.832$\pm$0.031 \\ W: 0.96$\pm$0.018\end{tabular} & \begin{tabular}{@{}c@{}}A: 0.944$\pm$0.009 \\ P: 0.906$\pm$0.014\\ W :0.984$\pm$0.005\end{tabular} & 5.5 \\
        \hline
        SRS & \cellcolor{green!10} 0.001 & 2.7 & 4m 2s  & \begin{tabular}{@{}c@{}}A: 0.931$\pm$0.013 \\ P:0.9$\pm$0.017 \\ W: 0.963$\pm$0.021\end{tabular} & \begin{tabular}{@{}c@{}}A: 0.886$\pm$0.013  \\ P: 0.815$\pm$0.016  \\ W: 0.957$\pm$0.01\end{tabular} & \cellcolor{green!10}\begin{tabular}{@{}c@{}}A: 0.937$\pm$0.005 \\ P: 0.906$\pm$0.008\\ W: 0.968$\pm$0.004\end{tabular} & 0+2.25 \\
        \hline
        DES & 0.24 & 3.21 & 3m 59s & \begin{tabular}{@{}c@{}}A: 0.894$\pm$0.022 \\ P: 0.824$\pm$0.04 \\ W: 0.964$\pm$0.03\end{tabular} & \begin{tabular}{@{}c@{}}A: 0.86$\pm$0.031  \\ P: 0.761$\pm$0.045 \\ W: 0.96$\pm$0.019\end{tabular} & \begin{tabular}{@{}c@{}}A: 0.919$\pm$0.006 \\ P: 0.855$\pm$0.009\\ W: 0.984$\pm$0.005\end{tabular} & 0.002+2.24 \\
        \hline
        NRMD & 0.09 & 2.3 & 3m 58s & \begin{tabular}{@{}c@{}}A: 0.901$\pm$0.016 \\ P: 0.887$\pm$0.021 \\ W: 0.914$\pm$0.023\end{tabular} & \begin{tabular}{@{}c@{}}A: 0.846$\pm$0.018 \\ P: 0.743$\pm$0.043 \\ W: 0.949$\pm$0.011\end{tabular} & \begin{tabular}{@{}c@{}}A: 0.908$\pm$0.009\\ P: 0.857$\pm$0.016\\ W: 0.958$\pm$0.003\end{tabular} & 0.001+2.23 \\
        \hline
        MMS & 0.05 & 2.3 & 4m & \cellcolor{green!10} \begin{tabular}{@{}c@{}}A: 0.935$\pm$0.012 \\ P: 0.927$\pm$0.017 \\ W: 0.943$\pm$0.015\end{tabular} & \begin{tabular}{@{}c@{}}A: 0.858$\pm$0.009\\ P: 0.776$\pm$0.002 \\ W: 0.94$\pm$0.01 \end{tabular} & \begin{tabular}{@{}c@{}}A: 0.922$\pm$0.01 \\ P: 0.885$\pm$0.019\\ W: 0.958$\pm$0.006\end{tabular} & 0.0004+2.23 \\ 
        \hline
        RKM & 1.22 & \cellcolor{green!10} 1.18 & 3m 52s & \begin{tabular}{@{}c@{}}A: 0.908$\pm$0.015\\ P: 0.821$\pm$0.029 \\ W: 0.995$\pm$0.006\end{tabular} & \begin{tabular}{@{}c@{}}A: 0.881$\pm$0.009  \\ P: 0.829$\pm$0.019 \\ W: 0.934$\pm$0.009\end{tabular} & \begin{tabular}{@{}c@{}}A: 0.927$\pm$0.007 \\ P: 0.881$\pm$0.015\\ W: 0.972$\pm$0.004\end{tabular} & \cellcolor{green!10} 0.005+2.18 \\
        \hline
        PRD & 0.38 & 2.68 & 4m & \begin{tabular}{@{}c@{}}A: 0.895$\pm$0.024\\ P: 0.849$\pm$0.038\\ W: 0.941$\pm$0.02\end{tabular} & \begin{tabular}{@{}c@{}}A: 0.872$\pm$0.033  \\ P: 0.793$\pm$0.066  \\ W: 0.951$\pm$0.011\end{tabular} & \begin{tabular}{@{}c@{}}A: 0.916$\pm$0.013\\ P: 0.856$\pm$0.022\\ W: 0.967$\pm$0.014\end{tabular} & 0.001+2.22 \\
        \hline
        PHL & 0.67 & 3.74 & 3m 59s & \begin{tabular}{@{}c@{}}A: 0.897$\pm$0.02\\ P: 0.816$\pm$0.034\\ W: 0.977$\pm$0.017\end{tabular} & \cellcolor{green!10} \begin{tabular}{@{}c@{}}A: 0.887$\pm$0.016  \\ P: 0.834$\pm$0.023 \\ W: 0.941$\pm$0.016\end{tabular} & \begin{tabular}{@{}c@{}}A: 0.923$\pm$0.004 \\ P: 0.877$\pm$0.007\\ W: 0.97$\pm$0.004\end{tabular} & 0.004+2.23 \\
        \hline
        FES & 8.63 & 3.12 & 3m 55s & \begin{tabular}{@{}c@{}}A: 0.884$\pm$0.002 \\ P: 0.824$\pm$0.033  \\ W: 0.945$\pm$0.016\end{tabular} & \begin{tabular}{@{}c@{}}A: 0.874$\pm$0.014  \\ P: 0.796$\pm$0.024  \\ W: 0.951$\pm$0.009\end{tabular} & \begin{tabular}{@{}c@{}}A: 0.911$\pm$0.008 \\ P: 0.854$\pm$0.013 \\ W: 0.968$\pm$0.004 \end{tabular} &  0.08+2.17 \\
        \hline
    \end{tabular}}
    \caption[Table Results for Roboflow Dataset and 75\% Reduction Rate]{Table results for Roboflow dataset and 75\% reduction rate. The 'Precision', 'Recall' and '$mAP$@.5' columns display mean and standard deviation values for the specified variables. The '$\text{CO}_2(g)$' column indicates the grams of $\text{CO}_2$ emitted during the application of the reduction method and during the fine-tuning. The 'R Time(s)' column shows the time in seconds for data reduction, while the 'FT Time' column displays the time spent on fine-tuning the model. We have highlighted in red the values obtained during fine-tuning with the complete dataset,which serve as the reference. Additionally, in green, we highlight the best reduction method for each metric.}
    \label{tab:B1-75}
\end{table}

\paragraph*{Mobility Aid Dataset}

Initially, we observe the training benchmark results with the complete dataset comprising 10,961 instances. This yields commendable outcomes, such as a mean average precision of 0.93, with a narrow standard deviation of 0.003 across all classes. Notably, the performance stands out in the category of people in wheel\-chairs (W), surpassing the overall average, while performance in other categories is noteworthy. Other categories were not included in Table \ref{tab:B2-75} and Table \ref{tab:B2-90}.
In general, superior performance is evident for the wheelchair, push-wheelchair, and walking-frame categories compared to the pedestrian and crutches categories, which exhibit below-average performance.

Moving to Table~\ref{tab:B2-75}, which illustrates the results for a reduction rate of 75\%, we observe a significant reduction in training time, approximately 50\%. Reduction times are just seconds for most methods, extending to minutes for PRD, PHL and FES. $\text{CO}_2$ emissions also witness a substantial decrease, around 55\% for all methods, except for PRD, PHL and FES, which emit more $\text{CO}_2$ due to an extended computation time during data reduction. Finally,  across all methods, we managed to maintain the performance achieved with the complete training set. This underscores the practical significance of these methods in reducing computation time and consequently lowering $\text{CO}_2$ emissions during model fine-tuning. The exceptions are the RKM and NRMD methods, which exhibit a performance drop.

In Table~\ref{tab:B2-90}, we present similar findings, but this time with a reduction rate of 90\%. The primary observation is a decrease in performance across various metrics, including precision, recall, and mean average precision. Notably, while there is an overall performance loss due to reduced metrics in other categories, the decline in the wheelchair category is comparatively less pronounced. Despite this reduction in performance, we managed to cut down the training time to 43 minutes, representing about 67\% of the training time without dataset reduction. A corresponding decrease in $\text{CO}_2$ consumption is observed. Despite the general decline, certain methods, such as SRS, MMS, PRD, PHL and FES, demonstrate a relatively robust maintenance of performance.

\begin{figure}[H]
        \centering
        \includegraphics[width=\textwidth]{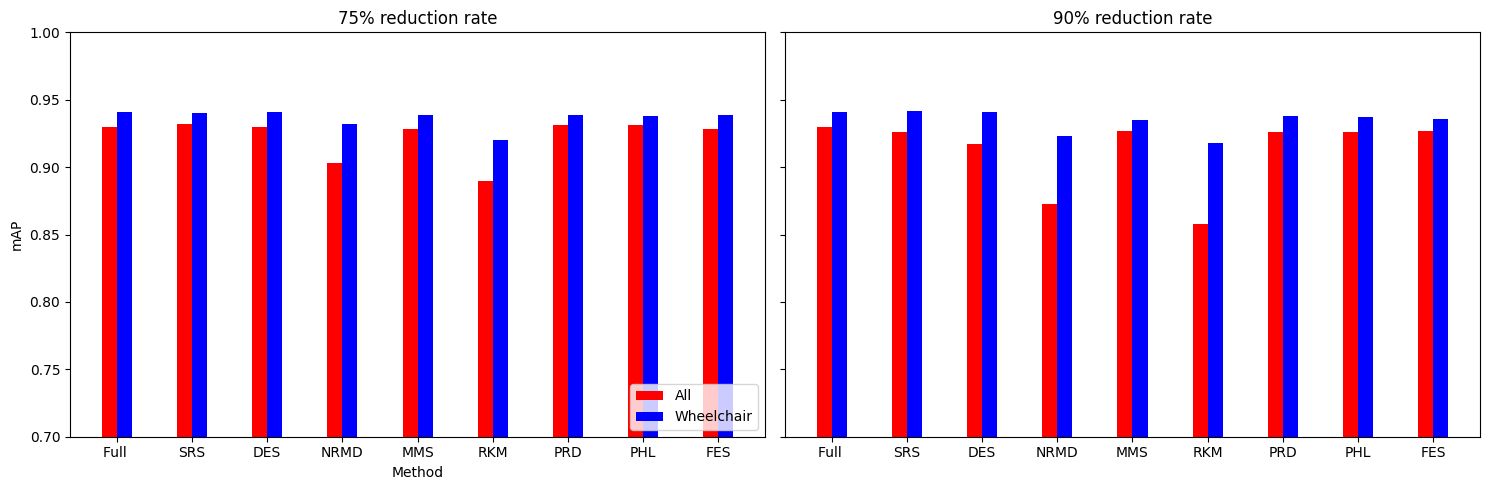}
        \caption[$mAP$ Values on Mobility Aid Dataset]{\label{$mAP$MobilityAid} $mAP$ values on Mobility Aid dataset when using a 75\% reduction rate(first column) and when using a 90\% of reduction rate (second column).}
\end{figure}

A visual representation of the mean $mAP$ values for wheel\-chairs, overall, and for each method alongside the full dataset is provided in Figure~\ref{$mAP$MobilityAid}. 
This visualization offers a clearer insight into the methods that yield optimal results. With a reduction rate of 75\%, we almost maintain performance in all methods, except NRMD and RKM. However, at a rate of reduction 90\%, some performance loss is evident, highlighting the efficacy of methods such as SRS, MMS, PRD, PHL and FES. In particular, the drop in performance for the wheelchair category is less pronounced compared to other categories. We only present the $mAP$ figure (Figure \ref{$mAP$MobilityAid}), as it serves as the most comprehensive performance metric for object detection evaluation.

With these results obtained for this dataset, we can confirm that the use of reduction methods within the proposed methodology, followed by fine-tuning YOLOv5 for object detection, led to a substantial reduction in both $\text{CO}_2$ emissions and computation time. Importantly, this reduction did not affect the model performance in object detection and localization.

\begin{table}[H]
    \centering
\resizebox{0.8\columnwidth}{!}{
    \begin{tabular}{|c|c|c|c|c|c|c|c|}
        \hline
        \cellcolor{gray!20}\textbf{Method}& \cellcolor{gray!20}\textbf{R Time(s)}& \cellcolor{gray!20}\textbf{$\varepsilon$}& \cellcolor{gray!20}\textbf{FT Time}& \cellcolor{gray!20}\textbf{Precision}& \cellcolor{gray!20}\textbf{Recall}& \cellcolor{gray!20}\textbf{$mAP$@.5}& \cellcolor{gray!20}\textbf{CO}$_2$\textbf{(g)}\\
        \hline
        \rowcolor{red!10}
        - & - & - & 2h 8m & \begin{tabular}{@{}c@{}}A: 0.91$\pm$0.008  \\ W: 0.994$\pm$0.001 \end{tabular} & \begin{tabular}{@{}c@{}}A: 0.875$\pm$0.007  \\ W: 0.896$\pm$0.01\end{tabular} & \begin{tabular}{@{}c@{}}A: 0.93$\pm$0.003 \\ W: 0.941$\pm$0.003\end{tabular} &  73.85 \\
        \hline
        SRS & \cellcolor{green!10} 0.006 & 0.86 & 1h 2m & \begin{tabular}{@{}c@{}}A: 0.912$\pm$0.007  \\ W: 0.979$\pm$0.029 \end{tabular} & \begin{tabular}{@{}c@{}}A: 0.876$\pm$0.007 \\ W: 0.887$\pm$0.011 \end{tabular} & \cellcolor{green!10} \begin{tabular}{@{}c@{}}A: 0.932$\pm$0.004  \\ W: 0.94$\pm$0.003\end{tabular} & 0+34.12 \\
        \hline
        DES & 7.98 & 0.87 & 1h 2m & \begin{tabular}{@{}c@{}}A: 0.91$\pm$0.009 \\ W: 0.989$\pm$0.005 \end{tabular} & \cellcolor{green!10} \begin{tabular}{@{}c@{}}A: 0.876$\pm$0.006  \\ W: 0.893$\pm$0.004\end{tabular} & \begin{tabular}{@{}c@{}}A: 0.93$\pm$0.006  \\ W: 0.941$\pm$0.001\end{tabular} &  0.04+34.1  \\
        \hline
        NRMD & 3.83 & 1.04 & 1h 2m & \begin{tabular}{@{}c@{}}A: 0.857$\pm$0.004  \\ W: 0.986$\pm$0.007 \end{tabular} & \begin{tabular}{@{}c@{}}A: 0.83$\pm$0.006 \\ W: 0.865$\pm$0.031\end{tabular} & \begin{tabular}{@{}c@{}}A: 0.903$\pm$0.004  \\ W: 0.932$\pm$0.002\end{tabular} &  \cellcolor{green!10} 0.015+33.94 \\
        \hline
        MMS & 5.88 & 0.66 & 1h 2m & \begin{tabular}{@{}c@{}}A: 0.911$\pm$0.006 \\ W: 0.994$\pm$0.003\end{tabular} & \begin{tabular}{@{}c@{}}A: 0.875$\pm$0.004  \\ W: 0.891$\pm$0.007 \end{tabular} & \begin{tabular}{@{}c@{}}A: 0.928$\pm$0.005 \\ W: 0.939$\pm$0.005\end{tabular} &  0.03+34.13\\
        \hline
        RKM & 31 & 0.74 & 1h 2m & \begin{tabular}{@{}c@{}}A: 0.845$\pm$0.013  \\ W: 0.961(0.008 \end{tabular} & \begin{tabular}{@{}c@{}}A: 0.795$\pm$0.006  \\ W: 0.871$\pm$0.008 \end{tabular} & \begin{tabular}{@{}c@{}}A: 0.89$\pm$0.006  \\ W: 0.92$\pm$0.005 \end{tabular} &  0.023+34.1 \\
        \hline
        PRD & 457 & \cellcolor{green!10} 0.59 & 1h 1m & \begin{tabular}{@{}c@{}}A: 0.912$\pm$0.011  \\ W: 0.989$\pm$0.005\end{tabular} & \begin{tabular}{@{}c@{}}A: 0.876$\pm$0.007  \\ W: 0.889$\pm$0.008\end{tabular} & \begin{tabular}{@{}c@{}}A: 0.931$\pm$0.004 \\ W: 0.939$\pm$0.002 \end{tabular} &  1.82+33.91 \\
        \hline
        PHL & 314 & 0.9 & 1h 3m & \begin{tabular}{@{}c@{}}A: 0.912$\pm$0.007 \\ W: 0.994$\pm$0.003\end{tabular} & \begin{tabular}{@{}c@{}}A: 0.875$\pm$0.009 \\ W: 0.896$\pm$0.003\end{tabular} & \begin{tabular}{@{}c@{}}A: 0.931$\pm$0.003 \\ W: 0.38$\pm$0.001 \end{tabular} &  1.79+35.09 \\
        \hline
        FES & 289 & 1 & 56m 33s & \cellcolor{green!10} \begin{tabular}{@{}c@{}}A: 0.912$\pm$0.004 \\ W: 0.993$\pm$0.001 \end{tabular} & \begin{tabular}{@{}c@{}}A: 0.875$\pm$0.005  \\ W: 0.887$\pm$0.009 \end{tabular} & \begin{tabular}{@{}c@{}}A: 0.928$\pm$0.004  \\ W: 0.939$\pm$0.003 \end{tabular} &  2.9+31.83 \\
        \hline
    \end{tabular}
    }
    \caption[Table Results for Mobility Aid Dataset and 75\% Reduction Rate]{Table results for Mobility Aid dataset and 75\% reduction rate. The 'Precision', 'Recall' and '$mAP$@.5' columns display mean and standard deviation values for the specified variables. The '$\text{CO}_2(g)$' column indicates the grams of $\text{CO}_2$ emitted during the application of the reduction method and during the fine-tuning. The 'R Time(s)' column shows the time in seconds for data reduction, while the 'FT Time' column displays the time spent on fine-tuning the model. We have highlighted in red the values obtained during fine-tuning with the complete dataset, which serve as the reference. Additionally, in green, we highlight the best reduction method for each metric.}
    \label{tab:B2-75}
\end{table}

\begin{table}[H]
    \centering
    \resizebox{0.8\columnwidth}{!}{
    \begin{tabular}{|c|c|c|c|c|c|c|c|}
        \hline
        \cellcolor{gray!20}\textbf{Method}& \cellcolor{gray!20}\textbf{R Time(s)}& \cellcolor{gray!20}\textbf{$\varepsilon$}& \cellcolor{gray!20}\textbf{FT Time}& \cellcolor{gray!20}\textbf{Precision}& \cellcolor{gray!20}\textbf{Recall}& \cellcolor{gray!20}\textbf{$mAP$@.5}& \cellcolor{gray!20}\textbf{CO}$_2$\textbf{(g)}\\
        \hline
        \rowcolor{red!10}
        - & - & - & 2h 8m & \begin{tabular}{@{}c@{}}A: 0.91$\pm$0.008  \\ W: 0.994$\pm$0.001 \end{tabular} & \begin{tabular}{@{}c@{}}A: 0.875$\pm$0.007  \\ W: 0.896$\pm$0.01\end{tabular} & \begin{tabular}{@{}c@{}}A: 0.93$\pm$0.003 \\ W: 0.941$\pm$0.003\end{tabular} &  73.85 \\
        \hline
        SRS & \cellcolor{green!10} 0.018 & 1.15 & 48m 31s & \begin{tabular}{@{}c@{}}A: 0.894$\pm$0.01  \\ W: 0.993$\pm$0.003\end{tabular} & \begin{tabular}{@{}c@{}}A: 0.867$\pm$0.007  \\ W: 0.886$\pm$0.015 \end{tabular} & \begin{tabular}{@{}c@{}}A: 0.926$\pm$0.002  \\ W: 0.942$\pm$0.002 \end{tabular} &  0+26.13 \\
        \hline
        DES & 8.13 & 1 & 43m 28s & \begin{tabular}{@{}c@{}}A: 0.875$\pm$0.014  \\ W: 0.971$\pm$0.02 \end{tabular} & \begin{tabular}{@{}c@{}}A: 0.862$\pm$0.007 \\ W: 0.893$\pm$0.011\end{tabular} & \begin{tabular}{@{}c@{}}A: 0.917$\pm$0.006 \\ W: 0.94$\pm$0.003 \end{tabular} &  0.04+24.06 \\
        \hline
        NRMD & 2.69 & 1.18 & 43m 17s & \begin{tabular}{@{}c@{}}A: 0.826$\pm$0.005  \\ W: 0.976$\pm$0.008\end{tabular} & \begin{tabular}{@{}c@{}}A: 0.773$\pm$0.02  \\ W: 0.846$\pm$0.02 \end{tabular} & \begin{tabular}{@{}c@{}}A: 0.873$\pm$0.009  \\ W: 0.923$\pm$0.005 \end{tabular} &  \cellcolor{green!10} 0.017+23.93 \\
        \hline
        MMS & 2.9 & \cellcolor{green!10} 0.84 & 43m 26s & \begin{tabular}{@{}c@{}}A: 0.904$\pm$0.006  \\ W: 0.996$\pm$0.001 \end{tabular} & \cellcolor{green!10} \begin{tabular}{@{}c@{}}A: 0.875$\pm$0.004  \\ W: 0.87$\pm$0.01\end{tabular} & \cellcolor{green!10} \begin{tabular}{@{}c@{}}A: 0.927$\pm$0.004 \\ W: 0.935$\pm$0.002 \end{tabular} &  0.017+24.06 \\
        \hline
        RKM & 22.92 & 0.95 & 43m 26s & \begin{tabular}{@{}c@{}}A: 0.812$\pm$0.008  \\ W: 0.968$\pm$0.007 \end{tabular} & \begin{tabular}{@{}c@{}}A: 0.767$\pm$0.012  \\ W: 0.855$\pm$0.024 \end{tabular} & \begin{tabular}{@{}c@{}}A: 0.858$\pm$0.004  \\ W: 0.918$\pm$0.003 \end{tabular} &  0.024+24.04 \\
        \hline
        PRD & 66.4 & 1 & 43m 28s & \cellcolor{green!10} \begin{tabular}{@{}c@{}}A: 0.908$\pm$0.01  \\ W: 0.99$\pm$0.006 \end{tabular} & \begin{tabular}{@{}c@{}}A: 0.868$\pm$0.007  \\ W: 0.883$\pm$0.01 \end{tabular} & \begin{tabular}{@{}c@{}}A: 0.926$\pm$0.007  \\ W: 0.938$\pm$0.004\end{tabular} &  0.29+24.05 \\
        \hline
        PHL & 317 & 1.04 & 43m  36s & \begin{tabular}{@{}c@{}}A: 0.898$\pm$0.006 \\ W: 0.989$\pm$0.005\end{tabular} & \begin{tabular}{@{}c@{}}A: 0.869$\pm$0.005 \\ W: 0.889$\pm$0.007 \end{tabular} & \begin{tabular}{@{}c@{}}A: 0.926$\pm$0.003 \\ W: 0.937$\pm$0.002\end{tabular} &  1.35+24.11 \\
        \hline
        FES & 269 & 1.46 & 43m 29s & \begin{tabular}{@{}c@{}}A: 0.906$\pm$0.007  \\ W: 0.992$\pm$0.005 \end{tabular} & \begin{tabular}{@{}c@{}}A: 0.87$\pm$0.005  \\ W: 0.875$\pm$0.033 \end{tabular} & \begin{tabular}{@{}c@{}}A: 0.927$\pm$0.003 \\ W: 0.936$\pm$0.002 \end{tabular} &  2.69+24.02 \\
        \hline        
    \end{tabular}}
    \caption[Table Results for Mobility Aid Dataset and 90\% Reduction Rate]{Table results for Mobility Aid dataset and 90\% reduction rate. The 'Precision', 'Recall' and '$mAP$@.5' columns display mean and standard deviation values for the specified variables. The '$\text{CO}_2(g)$' column indicates the grams of $\text{CO}_2$ emitted during the application of the reduction method and during the fine-tuning. The 'R Time(s)' column shows the time in seconds for data reduction, while the 'FT Time' column displays the time spent on fine-tuning the model. We have highlighted in red the values obtained during fine-tuning with the complete dataset, which serves as the reference. Additionally, in green, we highlight the best reduction method for each metric.}
    \label{tab:B2-90}
\end{table}

\section*{Data Availability}

\begin{itemize}
\item Collision: It consists of predicting whether a platoon
of vehicles will collide based on features such as the number of cars and their speed. The dataset consists of 107,210 examples with 25 numerical features and 2 classes; collision = 1 and collision = 0. This dataset is available at 
\\
\url{https://github.com/Cimagroup/Experiments-SurveyGreenAI}, 
\\
\url{https://doi.org/10.5281/zenodo.10844476}. 
\\
Apache 2.0 license.

\item Dry Bean: This dataset was created by taking pictures of dry beans from 7 different types and calculating some geometric features from the images, such as the area, the perimeter and the eccentricity.  consists of predicting the type of dry bean based on these geometric features. The dataset contains 13,611 examples with 16 features and 7 classes. This dataset is available at 
\\
\url{https://doi.org/10.24432/C50S4B}. 
\\
CC BY 4.0 license.

\item Roboflow: It comprises 514 RGB images, each 416 pixels in both height and width. These images feature pedestrians and people in wheelchairs. The training dataset comprises 463 RGB images, in which a total of 499 pedestrians (annotated as P) and 616 wheelchair users (annotated as W) appear. The test dataset is composed of 51 RGB images, in which a total of 55 pedestrians and 65 wheelchair users appear. This dataset is available at \\
\url{https://universe.roboflow.com/2458761304-qq-com/wheelchair-detection}. \\
CC BY 4.0 license.

\item Mobility Aid: The dataset Mobility Aid 2 [61] is composed of 17079 RGB images, 10961 are part of the training dataset and the 6118 remaining are part of the test dataset. In this dataset we can find five types of objects: pedestrians (8371), wheelchair users (6458), person pushing a wheelchair (3323), person with crutches (5374) and person with a walking-frame (7649). This dataset is available at \\
\url{http://mobility-aids.informatik.uni-freiburg.de/}.

\end{itemize}

\section*{Code Availability}

Source code is available at 
\\
\url{https://github.com/Cimagroup/SurveyGreenAI}
\\
\url{https://github.com/Cimagroup/Experiments-SurveyGreenAI}
\\
Apache 2.0 license.

\section*{Acknowledgements}
The work was supported in part by the European Union HORIZON-CL4-2021-HUMAN-01-01 under grant agreement 101070028 (REXASI-PRO).
\clearpage

\addcontentsline{toc}{section}{References}
\bibliographystyle{plainurl}
\bibliography{biblio}
\end{document}